\documentclass{sigkddExp}

\usepackage{graphicx}
\usepackage{times}
\usepackage{latexsym}
\usepackage{comment}
\usepackage{amsmath}
\usepackage{xcolor}
\usepackage{svg}
\usepackage{colortbl}
\usepackage{url}

\begin{document}
%
% --- Author Metadata here ---
% -- Can be completely blank or contain 'commented' information like this...
%\conferenceinfo{WOODSTOCK}{'97 El Paso, Texas USA} % If you happen to know the conference location etc.
%\CopyrightYear{2001} % Allows a non-default  copyright year  to be 'entered' - IF NEED BE.
%\crdata{0-12345-67-8/90/01}  % Allows non-default copyright data to be 'entered' - IF NEED BE.
% --- End of author Metadata ---

\title{Context-Aware Counterfactual Data Augmentation for Gender Bias Mitigation in
Language Models}
%\subtitle{[Extended Abstract]
% You need the command \numberofauthors to handle the "boxing"
% and alignment of the authors under the title, and to add
% a section for authors number 4 through n.
%
% Up to the first three authors are aligned under the title;
% use the \alignauthor commands below to handle those names
% and affiliations. Add names, affiliations, addresses for
% additional authors as the argument to \additionalauthors;
% these will be set for you without further effort on your
% part as the last section in the body of your article BEFORE
% References or any Appendices.

\numberofauthors{3}
%
% You can go ahead and credit authors number 4+ here;
% their names will appear in a section called
% "Additional Authors" just before the Appendices
% (if there are any) or Bibliography (if there
% aren't)

% Put no more than the first THREE authors in the \author command
%%You are free to format the authors in alternate ways if you have more 
%%than three authors.

\author{
%% e-mail address with \email.
\alignauthor Shweta Parihar \\
       \affaddr{University of Illinois at Chicago}\\
       \email{spari@uic.edu}
\alignauthor Liu Guangliang\\
       \affaddr{Michigan State University}\\
       \email{liuguan5@msu.edu	} \\
\and   
\alignauthor Natalie Parde\\
       \affaddr{University of Illinois at Chicago}\\
       \email{parde@uic.edu}
       \alignauthor Lu	Cheng\\
       \affaddr{University of Illinois at Chicago}\\
       \email{lucheng@uic.edu}   
}
% \additionalauthors{Additional authors: John Smith (The Th{\o}rvald Group,
% email: {\texttt{jsmith@affiliation.org}}) and Julius P.~Kumquat
% (The Kumquat Consortium, email: {\texttt{jpkumquat@consortium.net}}).}
% \date{30 July 1999}
\maketitle

\begin{abstract}
A challenge in mitigating social bias in fine-tuned language models (LMs) is the potential reduction in language modeling capability, which can harm downstream performance. Counterfactual data augmentation (CDA), a widely used method for fine-tuning, highlights this issue by generating synthetic data that may align poorly with real-world distributions or creating overly simplistic counterfactuals that ignore the social context of altered sensitive attributes (e.g., gender) in the pretraining corpus. To address these limitations, we propose a simple yet effective context-augmented CDA method, \textit{Context-CDA}, which uses large LMs to enhance the diversity and contextual relevance of the debiasing corpus. By minimizing discrepancies between the debiasing corpus and pretraining data through augmented context, this approach ensures better alignment, enhancing language modeling capability. We then employ uncertainty-based filtering to exclude generated counterfactuals considered low-quality by the target smaller LMs (i.e., LMs to be debiased), further improving the fine-tuning corpus quality. Experimental results on gender bias benchmarks demonstrate that \textit{Context}-CDA effectively mitigates bias without sacrificing language modeling performance while offering insights into social biases by analyzing distribution shifts in next-token generation probabilities.
\end{abstract}

\section{Introduction}
\vspace{3ex}
Language models (LMs) have achieved remarkable success in generating human-like text across a wide range of applications, from chatbots \cite{miller2017parlai} to translation \cite{zhu2023multilingual}. However, these models often inherit and amplify biases present in their training data, which can lead to outputs that reinforce stereotypes or perpetuate harmful prejudices. These biases partly stem from the massive datasets used to train LMs, which frequently reflect societal imbalances, discriminatory language, and historical injustices.

\par\vspace{0.3em}
Despite promising results in mitigating bias in LMs, a fundamental drawback of current debiasing methods is their potential to harm the core modeling abilities of LMs \cite{gallegos2024bias}. These methods often rely on strategies such as removing or altering biased data \cite{zhao2018gender} or introducing controlled outputs \cite{sheng2019woman}, which can reduce the model's exposure to natural linguistic patterns. By filtering or distorting the underlying data, these techniques can hinder the model’s ability to grasp the subtleties of language, leading to less fluent and contextually inaccurate text generation. This trade-off underscores the challenge of reducing bias while maintaining language proficiency. Take Counterfactual Data Augmentation (CDA) \cite{lu2020gender, zmigrod2019counterfactual} for gender bias as an example. CDA works by altering gender attributes to generate counterfactual examples, which are then used during fine-tuning to reduce bias in model predictions. While CDA is effective in reducing bias, it often leads to a degradation in LMs' language modeling ability  \cite{gallegos2024bias, fatemi2021improving, raza2024mbias}. A major reason for this is the synthetic data generated by CDA may not align well with real-world data distributions and can create overly simplistic counterfactuals that fail to account for the social context of altered sensitive attributes (e.g., gender) embedded in the pre-training corpus.

\par
\begin{figure*}[t]
\begin{center}
  \includegraphics[width=0.75\linewidth]{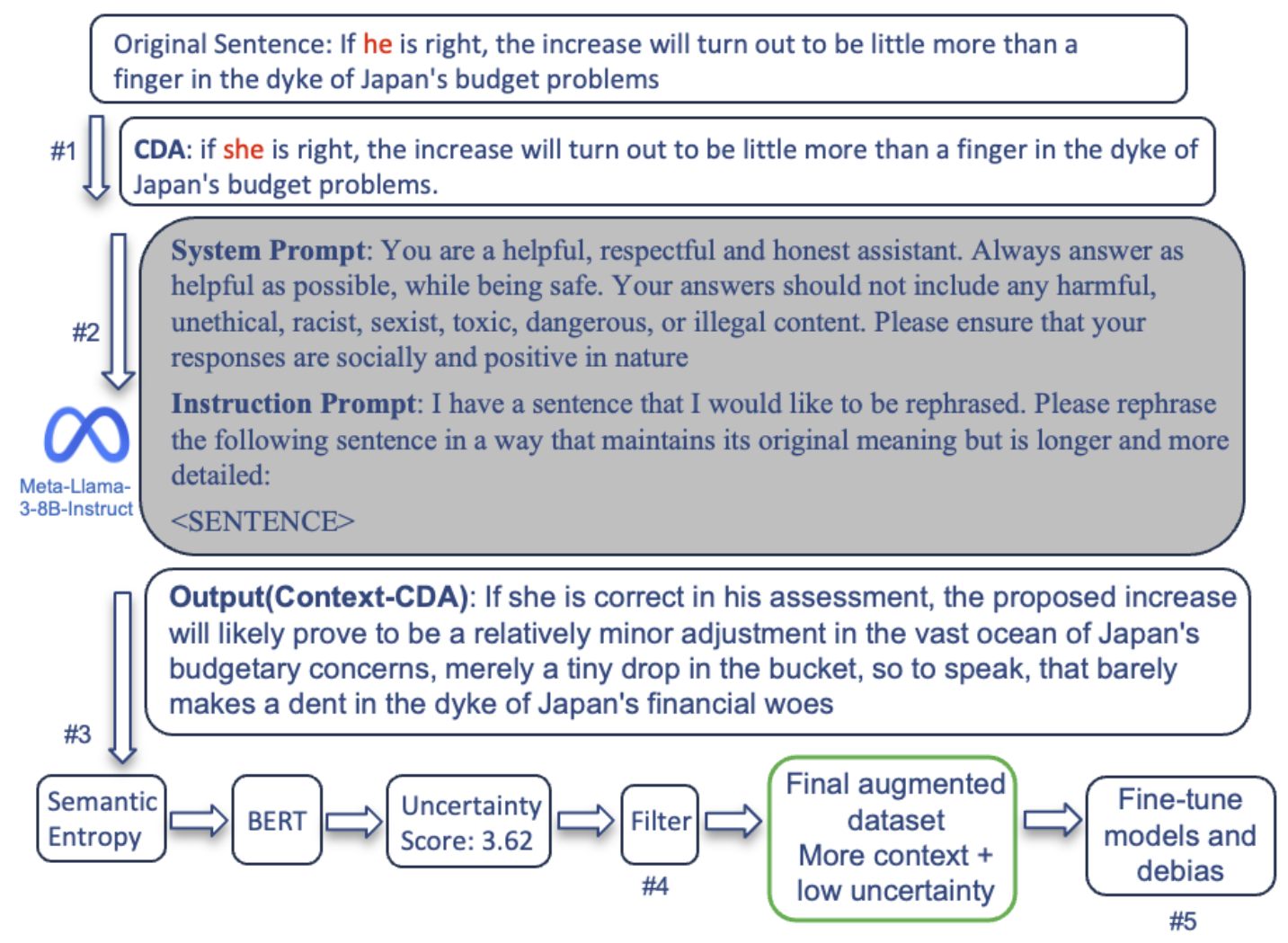}
\end{center}
\caption{An illustration of the proposed \textit{Context-CDA} pipeline using the StereoSet benchmark \protect\cite{nadeem2020stereoset} with BERT \protect\cite{devlin2019bert}. Step 1: Flip the gender words. Step 2: Use a larger LM (e.g., Llama-3-8B-Instruct \protect\cite{grattafiori2024llama}) with the system and instruction prompt to get the augmented data. Step 3: Use the target small LM (e.g., BERT) to calculate the semantic entropy of augmented data. Step 4: Filter the counterfactuals based on the semantic entropy. Step 5: Debias the target small LM.}
\label{Context-CDA-Pipeline}
\vspace{4mm}
\end{figure*}

\vspace{0.3em}
To overcome these challenges, we propose a simple yet effective approach, \textit{Context-CDA}, that leverages the generative capabilities of large LMs to augment the context used for debiasing. Particularly, larger LMs trained on vast and diverse corpora can generate context-rich counterfactual examples that are not only more diverse but also closely resemble natural language patterns (see an example in Figure \ref{Context-CDA-Pipeline}). This context-aware data augmentation process helps avoid the severe distortions often introduced by traditional CDA techniques, where overly simplified examples fail to consider the social context of altered sensitive attributes. Our approach thus enables a more comprehensive debiasing process, enhancing fairness while also preserving the language modeling abilities of LMs.

\par\vspace{0.3em}
However, the smaller LMs to be debiased often struggle to learn from text generated by larger LMs. Samples deemed challenging for the target LMs may hurt their language modeling capabilities \cite{huang2020counterfactually}. To improve the performance of the model and reduce the impact of low-quality generations, we further propose using an uncertainty-based filtering strategy. Particularly, we leverage semantic entropy \cite{kuhn2023semantic} to quantify the uncertainty of each augmented sample in our corpus. Semantic entropy advances other uncertainty estimation approaches (e.g., likelihoods) as it incorporates linguistic invariances created by shared meanings. High semantic entropy indicates that the target LM finds the sentence uncertain, ambiguous, or difficult to interpret due to multiple possible meanings or unpredictable word choices. We systematically filter out the counterfactual generations with the largest semantic entropy. Therefore, during fairness fine-tuning, the target LMs are trained on text that is clearer and less prone to misinterpretation. This filtering also ensures that the model can focus on learning meaningful and unambiguous patterns from the data, enhancing its ability to generalize and perform well in real-world applications. 

\par\vspace{0.3em}
Our contributions are as follows:
\textbf{Method}: We develop \textit{Context-CDA}, a new approach that produces context-rich, gender-flipped sentences. We enhance data quality for target LMs by using semantic entropy filtering to exclude ambiguous and irrelevant counterfactuals for fine-tuning. \textbf{Experiments}: Our extensive evaluations show that \textit{Context-CDA} preserves language modeling ability while reducing bias more effectively than traditional CDA methods (which may compromise on fluency). Semantic entropy-based filtering of low-quality counterfactuals further enhances debiasing effectiveness and maintains language fluency. Analysis of next-token distribution indicates a better gender balance and token diversity, leading to reduced skew towards male tokens, and increasing robustness. Comprehensive evaluation across five diverse model architectures, including BERT and DistilBERT (encoder-only models), T5 (encoder-decoder generative model), GPT-2, and Llama-3-1B (causal decoder-only generative model), demonstrates that these findings are consistent and robust across both discriminative and generative systems, validating the true model-agnostic nature of \textit{Context-CDA}.

% Reframe the contribution: even modest improvements matter given CDA’s known degradation of fluency. 

\begin{figure*}[t]
% \begin{center}
\centering
\begin{minipage}{0.48\linewidth}
    \centering
    \includegraphics[width=1\linewidth]{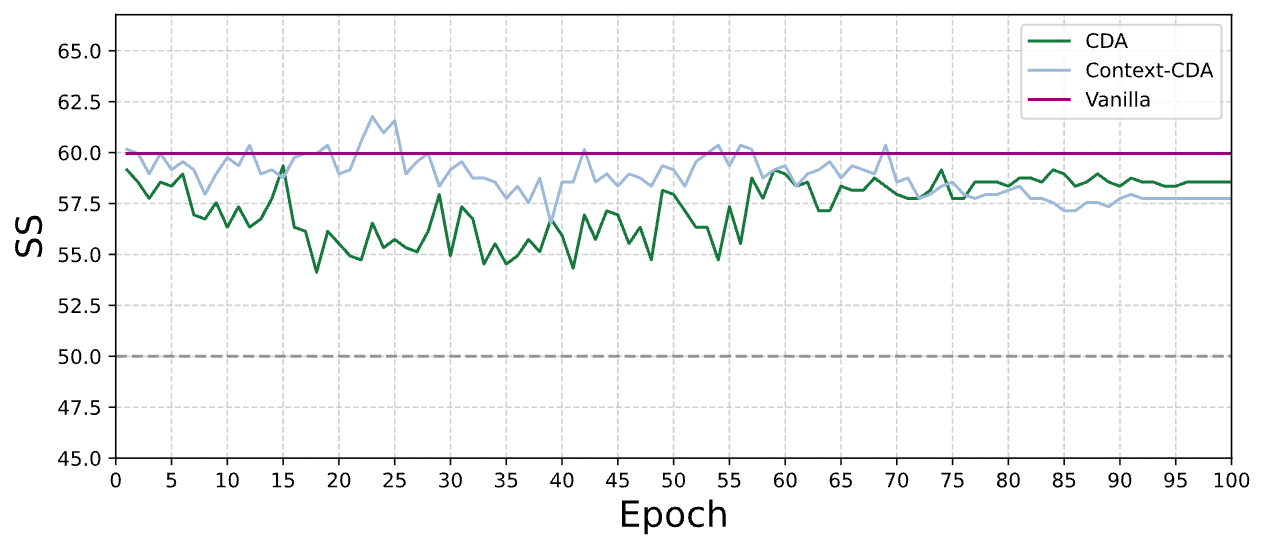}
    \caption{StereoSet bias score for BERT (Intrinsic bias). 50 indicates no bias.}
    \label{bert-ss}
\end{minipage}
\hfill
\begin{minipage}{0.48\linewidth}
    \centering
    \includegraphics[width=1\linewidth]{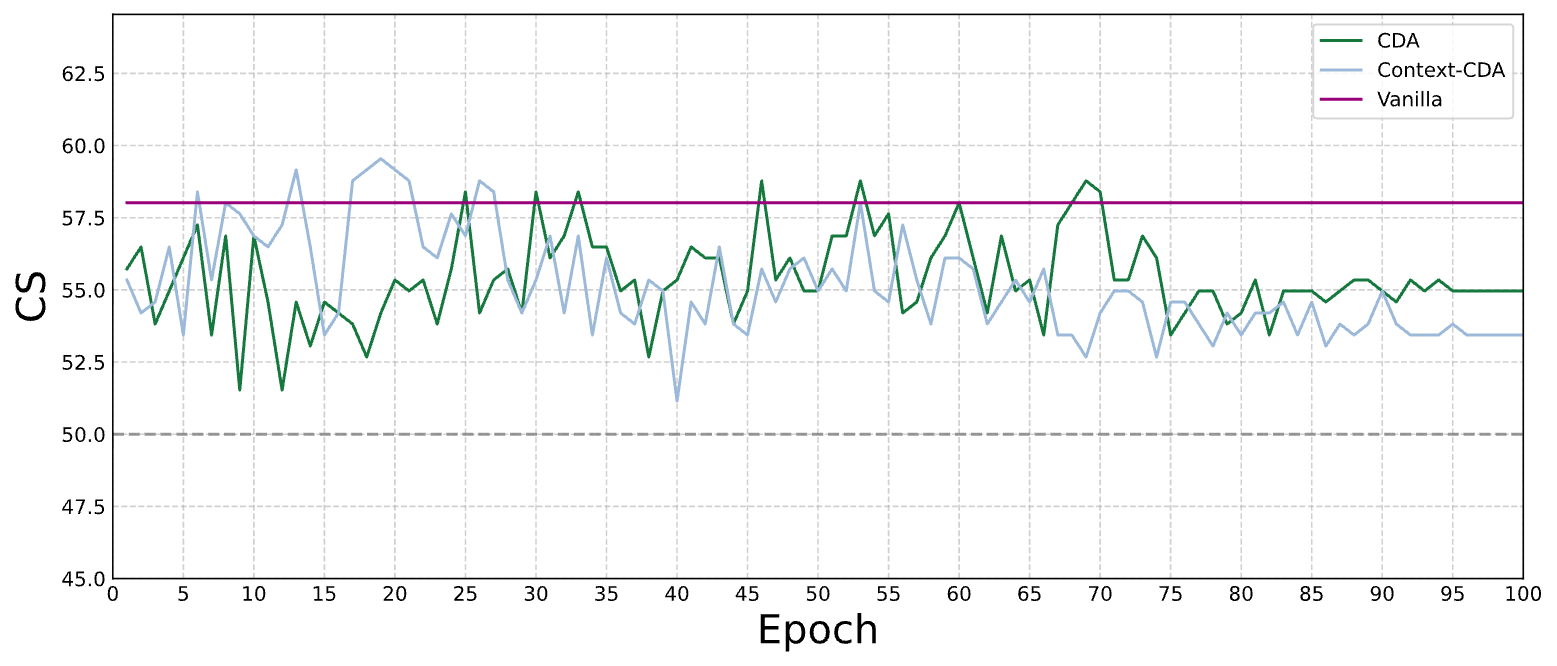}
    \caption{CrowS-Pairs bias score for BERT (Intrinsic bias). 50 indicates no bias.}
    \label{bert-cs}
\end{minipage}
% \end{center}

\begin{minipage}{0.48\linewidth}
    \centering
    \includegraphics[width=1\linewidth]{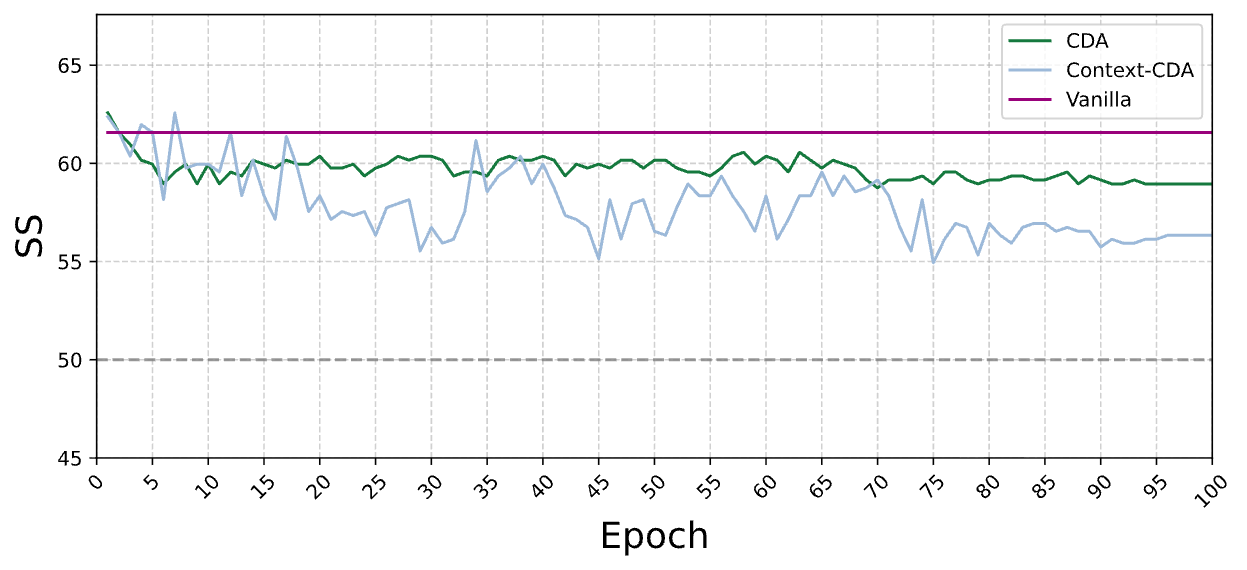}
    \caption{StereoSet bias score for DistilBERT.}
    \label{distilbert-ss}
\end{minipage}
\hfill
\begin{minipage}{0.48\linewidth}
    \centering
    \includegraphics[width=1\linewidth, height=3.6cm,
keepaspectratio=false]{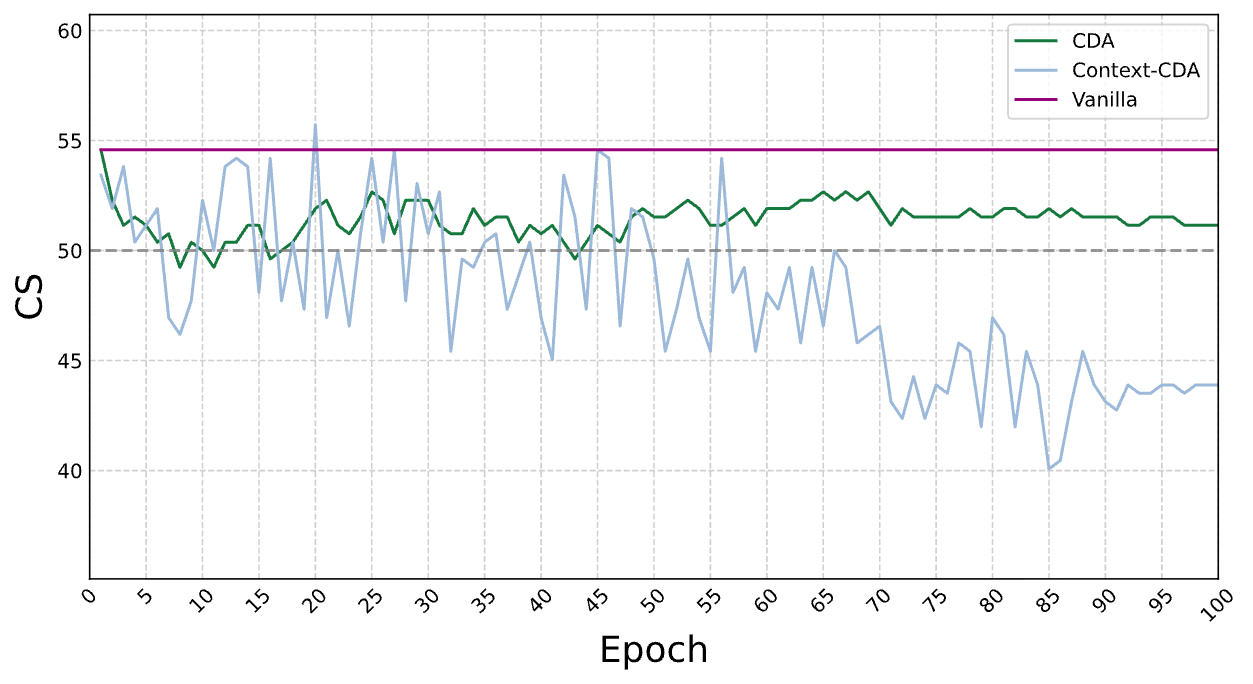}
    \caption{CrowS-Pairs bias score for DistilBERT.}
    \label{distilbert-cs}
\end{minipage}

\end{figure*}

\section{Related Work} 
\vspace{3ex}
An LM may be deployed in a different setting than that for which it was intended, such as with or without a human intermediary for automated decision-making \cite{suresh2021framework}. The primary causes of LM biases include inherent biases in the training data \cite{baumann2023bias}. Bias mitigation techniques are categorized by the different stages of LM workflow. Pre-processing mitigation techniques focus on reducing bias and unfairness early in the dataset or model inputs. One of the early formalizations of this approach involves CDA, which has emerged as a prominent technique for mitigating biases in language models by introducing minimally perturbed examples that alter specific attributes while preserving the overall semantics. The foundational work in CDA by \cite{zhao2018gender} demonstrated its effectiveness in reducing gender bias in coreference resolution by swapping gendered terms in training data. \cite{lu2020gender} extended this to a more general framework by applying CDA for neural NLP tasks by generating matched sentence pairs through gendered word interventions. \cite{zmigrod2019counterfactual} proposed a CDA method tailored for morphologically rich languages to generate grammatically correct gender-swapped sentences. 

\par\vspace{0.3em}
As described by \cite{webster2020measuring}, the CDA procedure can be applied in a one-sided manner (using only the counterfactual sentence for further training) or a two-sided manner (incorporating both the original and counterfactual sentences in the training data). \cite{maudslay2019s} introduce Counterfactual Data Substitution (CDS), a variant of CDA, where gendered terms in the training data are probabilistically replaced with their counterfactual counterparts, rather than duplicating each instance with a gender-swapped version. More recently, \cite{wu2021polyjuice} proposed Polyjuice, a controllable counterfactual generation system based on GPT-2, enabling diverse perturbation types for training and evaluation. \cite{qian2022perturbation} introduced a neural demographic perturber trained on a large human-annotated dataset (PANDA), and demonstrate that augmenting data via demographic perturbations improves model fairness with minimal performance trade-offs. Both works highlight the utility of automated, fine-grained counterfactual generation for robust and fair NLP. 

\par\vspace{0.3em}
Recent advances in CDA span diverse domains, reflecting its growing relevance in addressing data imbalance and spurious correlations. In NLP, \cite{tokpo2024fairflow} introduce FairFlow, a model-based CDA method that generates parallel counterfactuals without manual intervention, improving fairness while preserving fluency. Similarly, \cite{sreedhar2025novel} employ self-imitation reinforcement learning for CDA, emphasizing both fairness and robustness under contextual shifts. In sentiment analysis, \cite{wu2024novel} propose polarity-reversing augmentations using pre-trained transformers to enhance aspect-based sentiment classification. Beyond text, CDA has also seen traction in graph neural networks. For instance, CAGAD \cite{xiao2024counterfactual} utilizes diffusion models to generate counterfactual anomalies in graphs, improving anomaly detection in node representations. In reinforcement learning, ACAMDA \cite{sun2024acamda} recovers temporal causal structures to simulate realistic hypothetical scenarios for improved data efficiency. CAIAC \cite{urpi2024causal} swaps causally irrelevant state components to generate robust offline learning transitions. A parallel effort to improve interpretability can be seen in FCE-UTD \cite{li2024beyond}, which generates factor-level counterfactuals for causal explanations in Point-of-Interest (POI) recommendation. Together, these works underscore the versatility of CDA across modalities and its emerging role in mitigating bias.

\par\vspace{0.3em}
Additionally, CDA has been integrated with other debiasing methods such as adversarial training \cite{ravfogel2020null} and calibration strategies \cite{tan2019assessing} to improve robustness and generalization. However, concerns remain around the semantic validity and distributional shift introduced by counterfactuals, leading to research on controllable generation (e.g., using LMs or paraphrasing systems) and evaluation metrics to ensure linguistic and contextual coherence. Our work builds upon this by extending the CDA framework to include context-rich, gender-flipped sentences generated by a large LM following the work of \cite{han2024chatgpt} for debiasing along with semantic entropy filtering to exclude ambiguous counterfactuals and improve corpus quality.

\begin{figure*}[t]

% \begin{center}
\begin{minipage}{0.48\linewidth}
    \centering
    \includegraphics[width=1\linewidth]{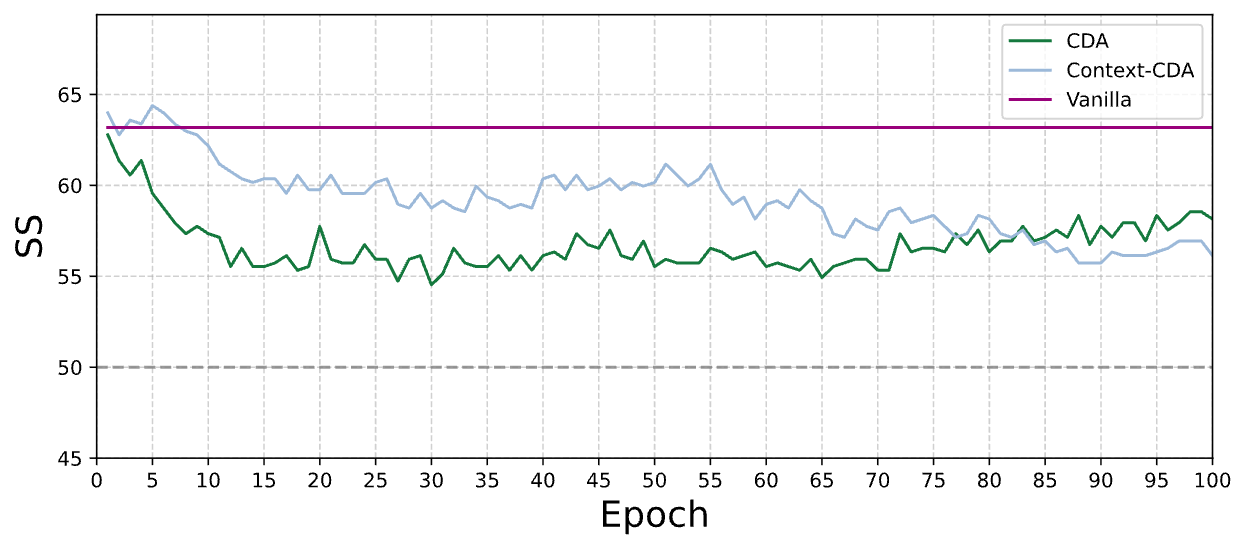}
    \caption{StereoSet bias score for GPT-2.}
    \label{gpt-ss}
\end{minipage}
\hfill
\begin{minipage}{0.48\linewidth}
    \centering
    \includegraphics[width=1\linewidth]{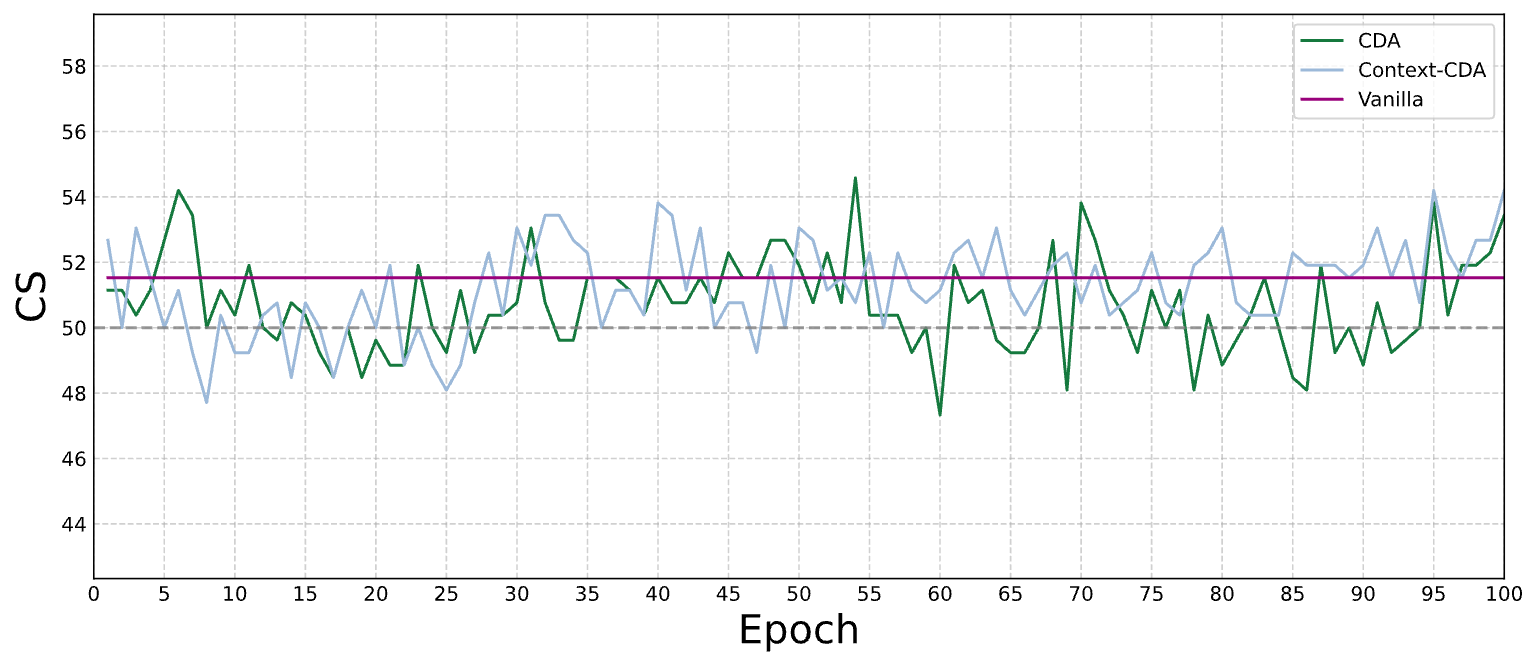}
    \caption{CrowS-Pairs bias score for GPT-2.}
    \label{gpt-cs}
\end{minipage}
% \end{center}

% \end{figure*}

% \begin{figure*}[t]

\begin{minipage}{0.48\linewidth}
    \centering
    \includegraphics[width=1\linewidth]{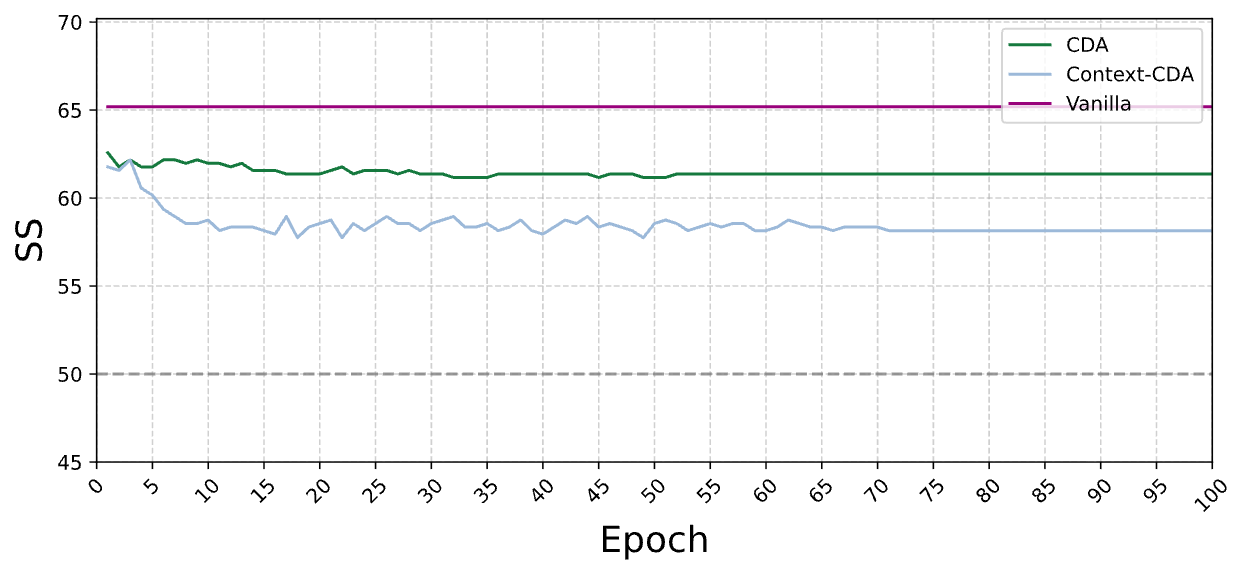}
    \caption{StereoSet bias score for Llama-3.2-1B.}
    \label{llama-ss}
\end{minipage}
\hfill
\begin{minipage}{0.48\linewidth}
    \centering
    \includegraphics[width=1\linewidth, height=3.6cm,
keepaspectratio=false]{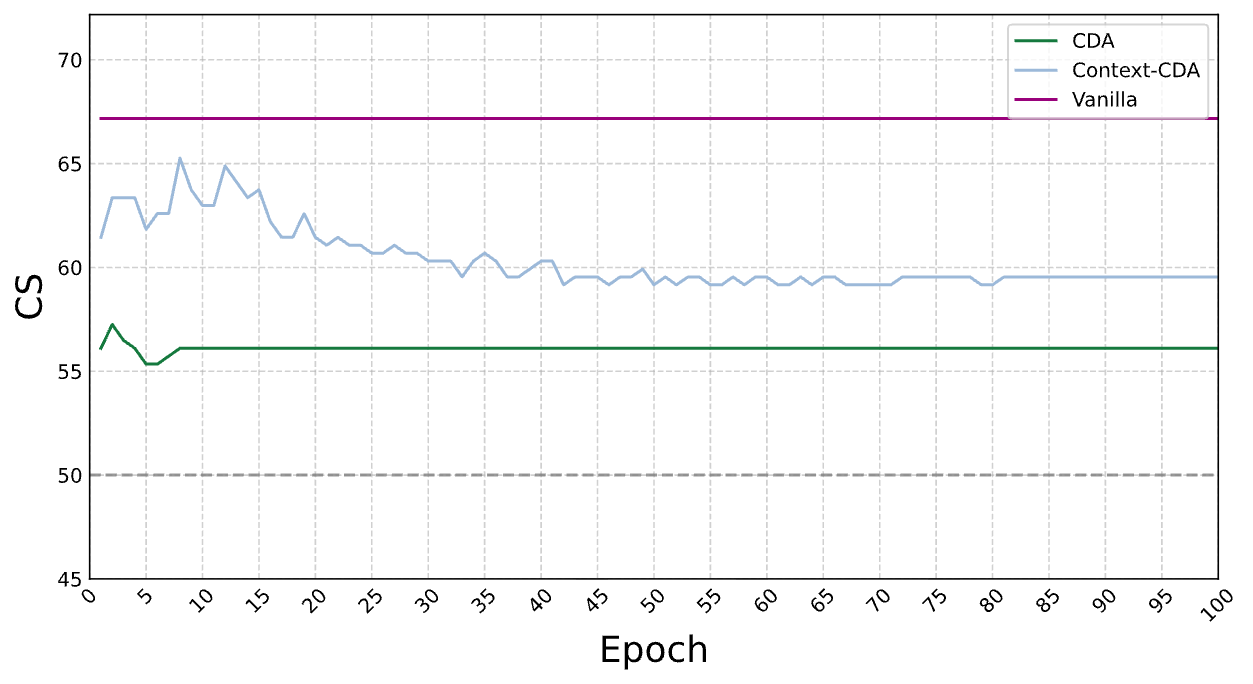}
    \caption{CrowS-Pairs bias score for Llama-3.2-1B.}
    \label{llama-cs}
\end{minipage}

\begin{minipage}{0.48\linewidth}
    \centering
    \includegraphics[width=1\linewidth]{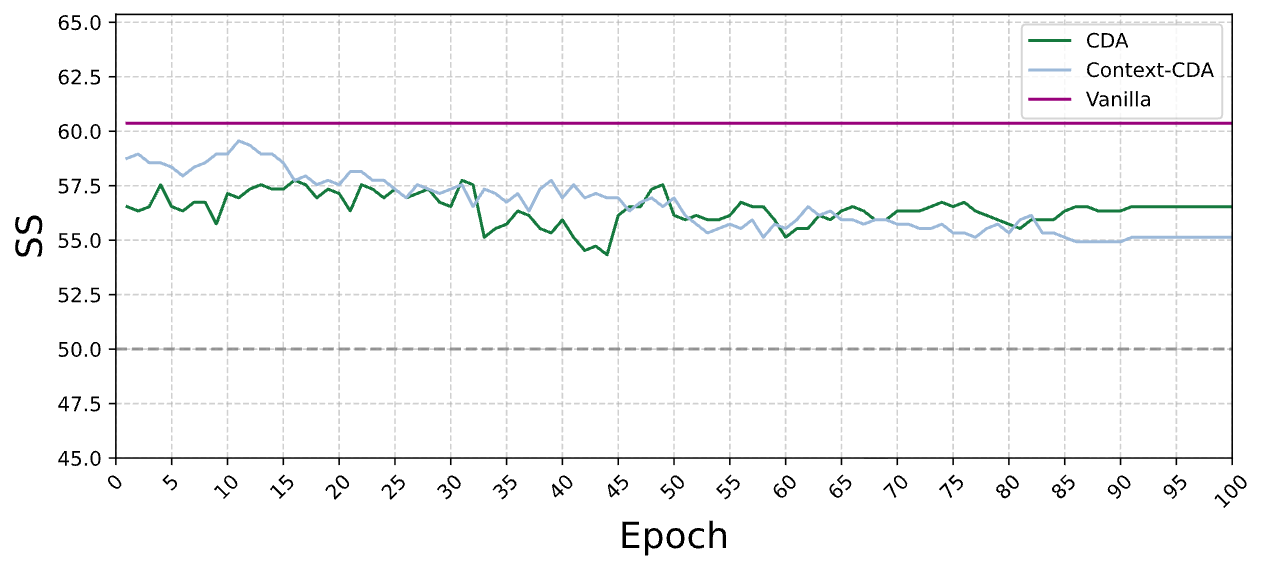}
    \caption{StereoSet bias score for T5.}
    \label{t5-ss}
\end{minipage}
\hfill
\begin{minipage}{0.48\linewidth}
    \centering
    \includegraphics[width=1\linewidth, height=3.6cm,
keepaspectratio=false]{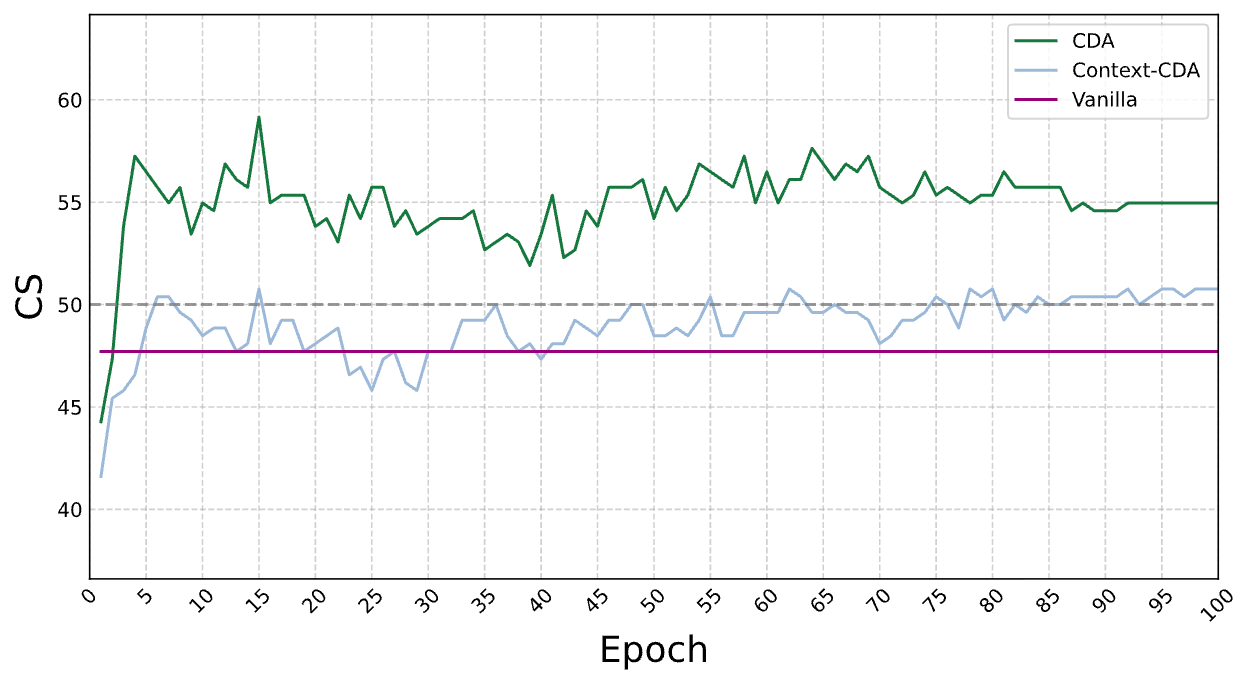}
    \caption{CrowS-Pairs bias score for T5.}
    \label{t5-cs}
\end{minipage}

\end{figure*}

\section{Methods}
\vspace{3ex}
The proposed method consists of three major steps: (1) Augmenting the context of CDA using large LMs, (2) Uncertainty-based filtering, and (3) Debiasing via fine-tuning on filtered counterfactuals. The overview of \textit{Context-CDA} is illustrated in Figure \ref{Context-CDA-Pipeline}.

\begin{figure*}[t]

\centering
% \begin{center}
\begin{minipage}{0.48\linewidth}
    \centering
    \includegraphics[width=1\linewidth]{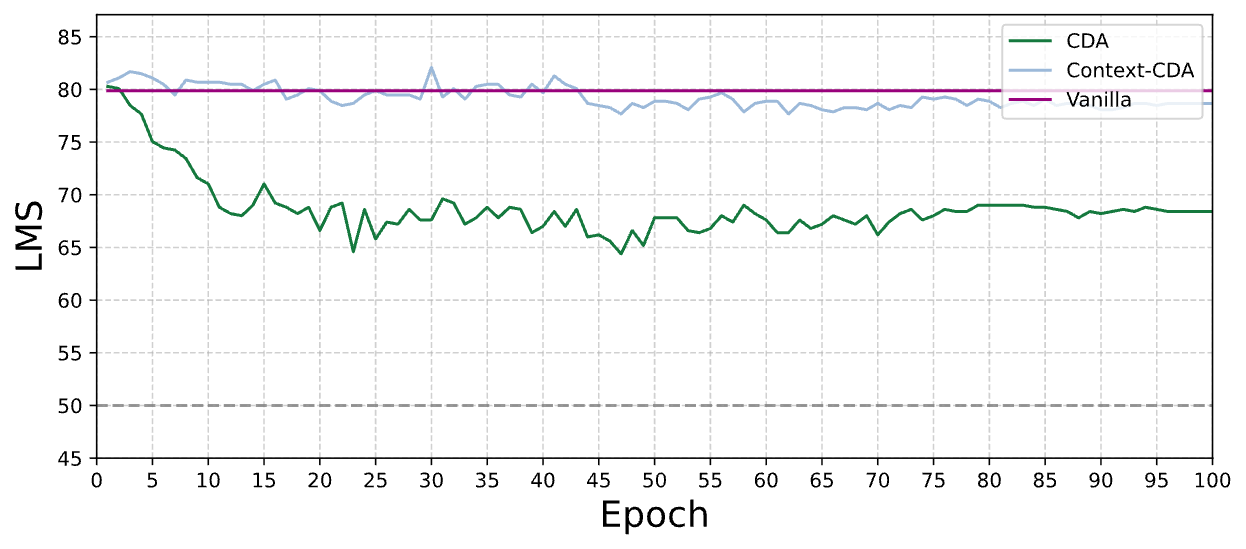}
    \caption{LMS score ($\uparrow$) for BERT. }
    \label{bert-lms}
\end{minipage}
\hfill
\begin{minipage}{0.48\linewidth}
    \centering
    \includegraphics[width=1\linewidth]{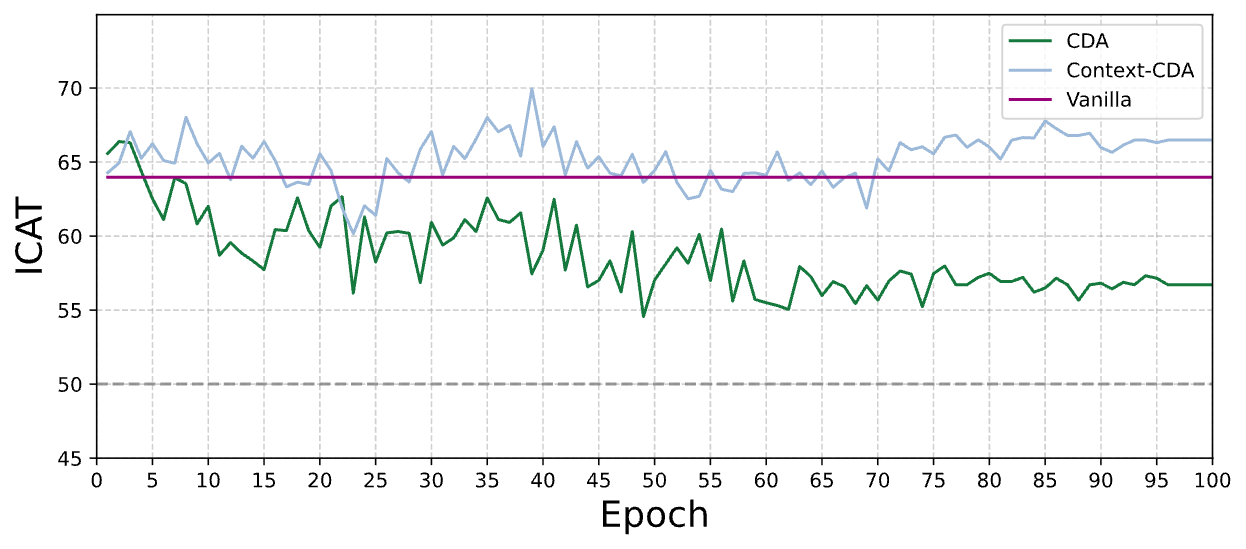}
    \caption{ICAT bias score ($\uparrow$) for BERT.}
    \label{bert-icat}
\end{minipage}
% \end{center}

% \begin{center}
\begin{minipage}{0.48\linewidth}
    \centering
    \includegraphics[width=1\linewidth]{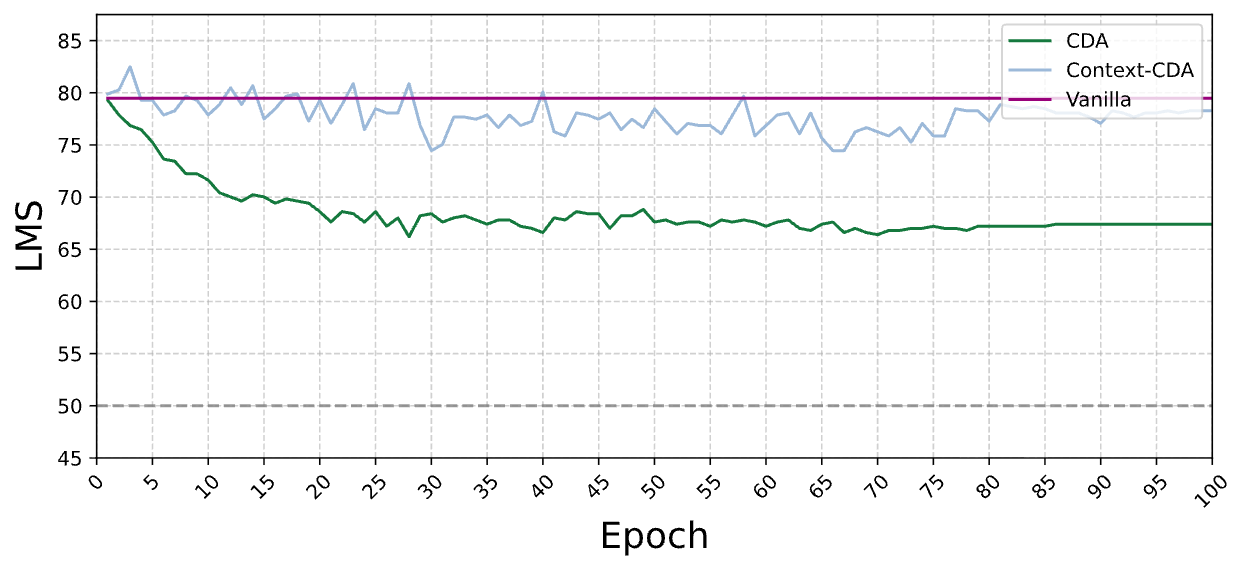}
    \caption{LMS score ($\uparrow$) for DistilBERT. }
    \label{distilbert-lms}
\end{minipage}
\hfill
\begin{minipage}{0.48\linewidth}
    \centering
    \includegraphics[width=1\linewidth]{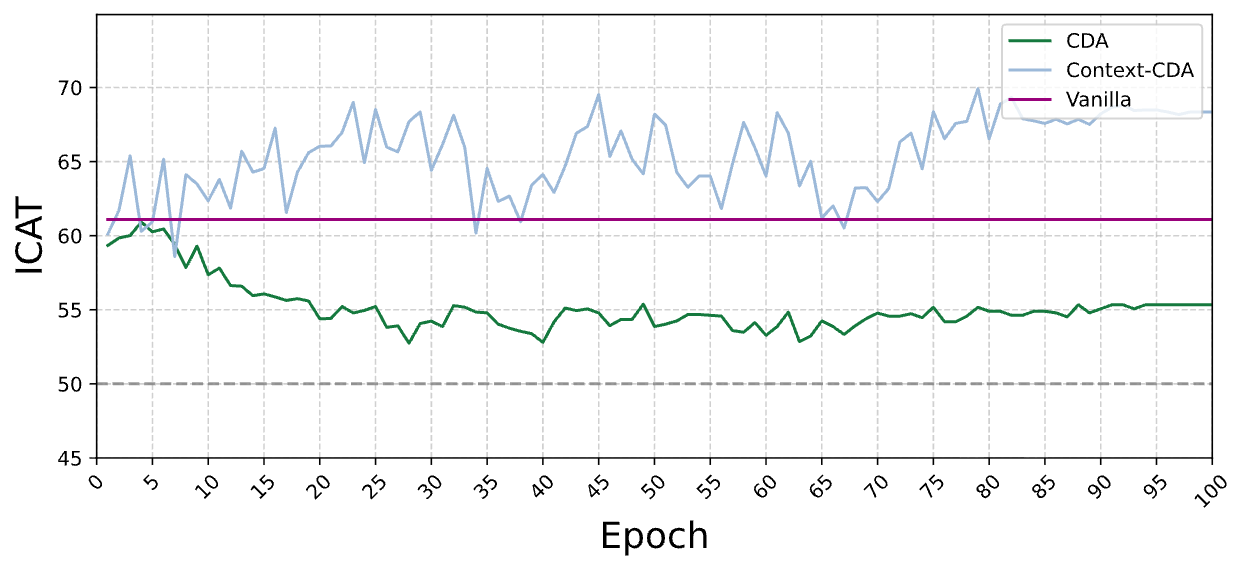}
    \caption{ICAT bias score ($\uparrow$) for DistilBERT.}
    \label{distilbert-icat}
\end{minipage}
% \end{center}

\end{figure*}

\subsection{Context-Aware CDA}
\vspace{3ex}
For gender bias, vanilla CDA alters gender-related words, which can degrade language modeling ability while being effective at bias mitigation \cite{gallegos2024bias,raza2024mbias}. One primary reason for this degradation is that the distribution of the corpus generated by vanilla CDA and then used for debiasing drifts from the distribution of the pre-trained corpus. When debiasing via fine-tuning the counterfactual generation, the n-gram distribution of the target LM is altered, resulting in the degradation of language modeling ability \cite{fatemi2021improving}. To address the limitation, we introduce \textit{Context-CDA}, a prompting-based method that advances traditional CDA by prompting a larger LM to generate contextually richer counterfactuals. Rather than merely flipping gender-related words, \textit{Context-CDA} refines the sentences to preserve their original meaning while providing a more natural and diverse context aligned with the LM’s pre-training distribution. This approach to context-aware data augmentation mitigates the severe distortions often caused by traditional CDA methods, where oversimplified examples fail to account for the social context of altered sensitive attributes. As a result, our method enables a more thorough debiasing process, promoting fairness while preserving the LMs' modeling capabilities. Specifically, after generating counterfactual examples using CDA \( \tilde{x}_i = \text{CDA}(x_i) \), we design a system prompt and an instruction prompt (illustrated in Figure \ref{Context-CDA-Pipeline}) to ask a Llama-3-8b-Instruct \cite{grattafiori2024llama} model to rephrase $\tilde{x}_i$ such that it contains more context. The resulting generation is denoted as $\tilde{x}^c_i$.

\begin{figure*}[t]

\centering

% \begin{center}
\begin{minipage}{0.48\linewidth}
    \centering
    \includegraphics[width=1\linewidth]{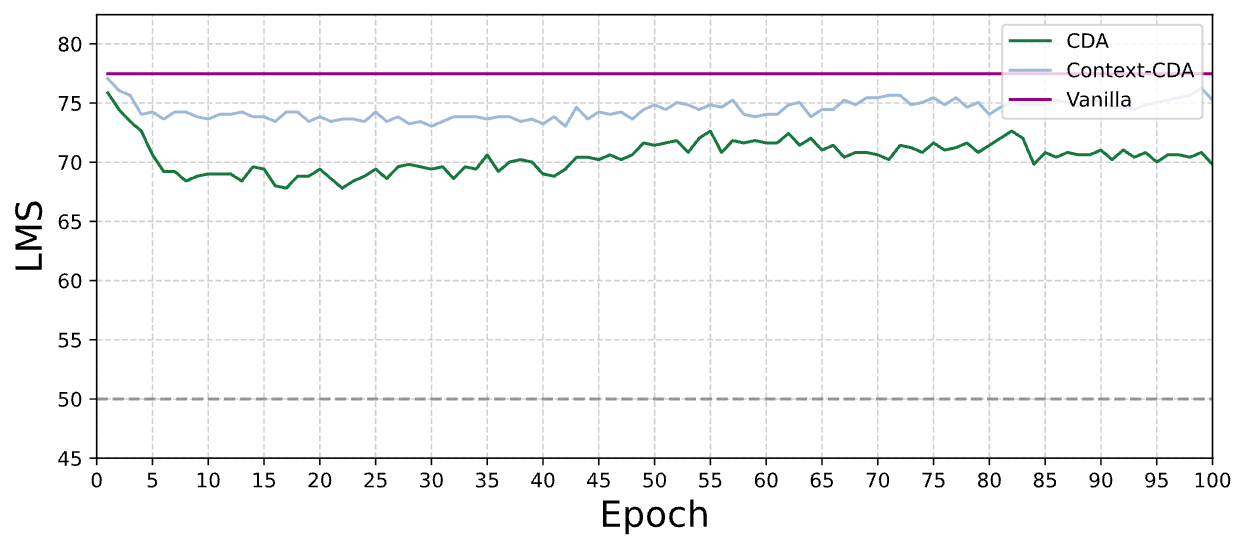}
    \caption{LMS score for GPT-2. }
    \label{gpt-lms}
\end{minipage}
\hfill
\begin{minipage}{0.48\linewidth}
    \centering
    \includegraphics[width=1\linewidth]{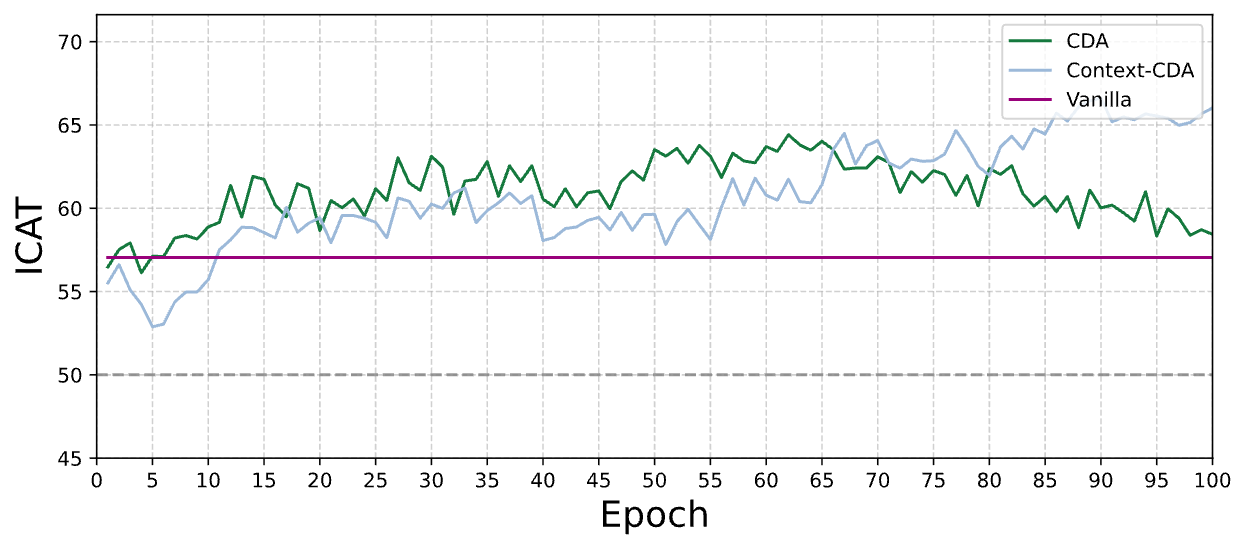}
    \caption{ICAT bias score for GPT-2.}
    \label{gpt-icat}
\end{minipage}
% \end{center}

% \begin{center}
\begin{minipage}{0.48\linewidth}
    \centering
    \includegraphics[width=1\linewidth]{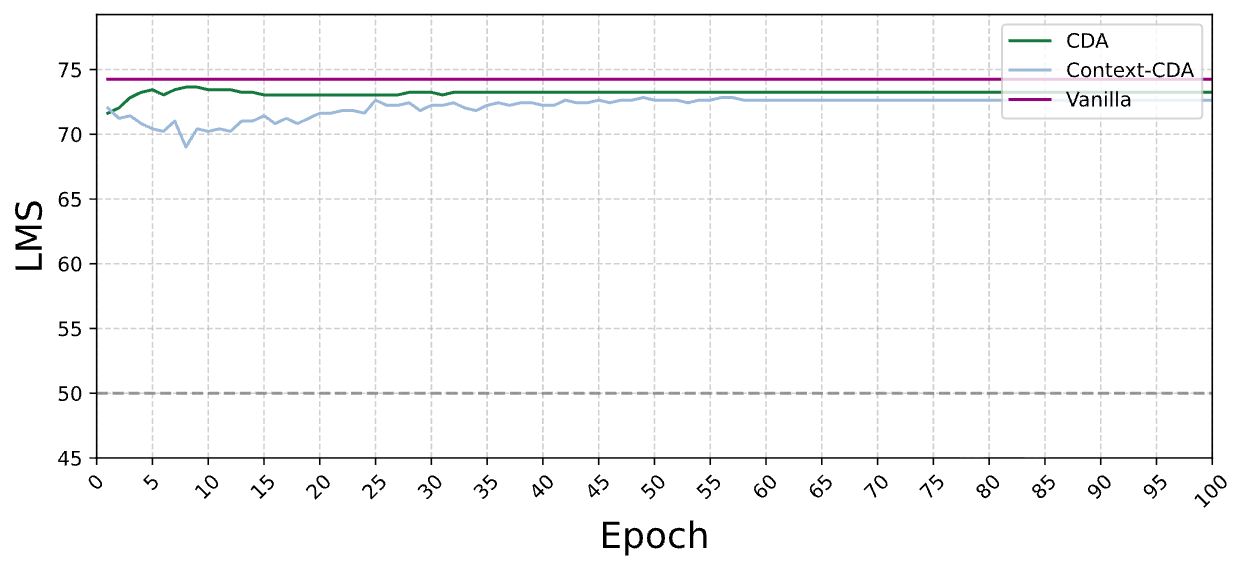}
    \caption{LMS score for Llama-3.2-1B. }
    \label{llama-lms}
\end{minipage}
\hfill
\begin{minipage}{0.48\linewidth}
    \centering
    \includegraphics[width=1\linewidth]{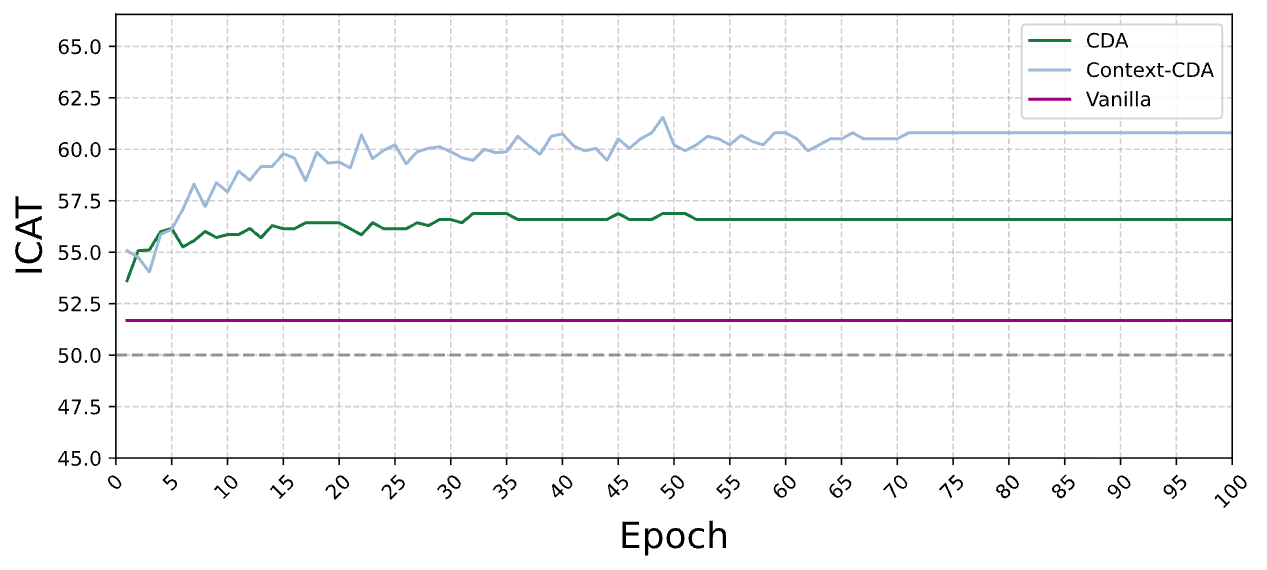}
    \caption{ICAT bias score for Llama-3.2-1B.}
    \label{llama-icat}
\end{minipage}
% \end{center}

% \begin{center}
\begin{minipage}{0.48\linewidth}
    \centering
    \includegraphics[width=1\linewidth]{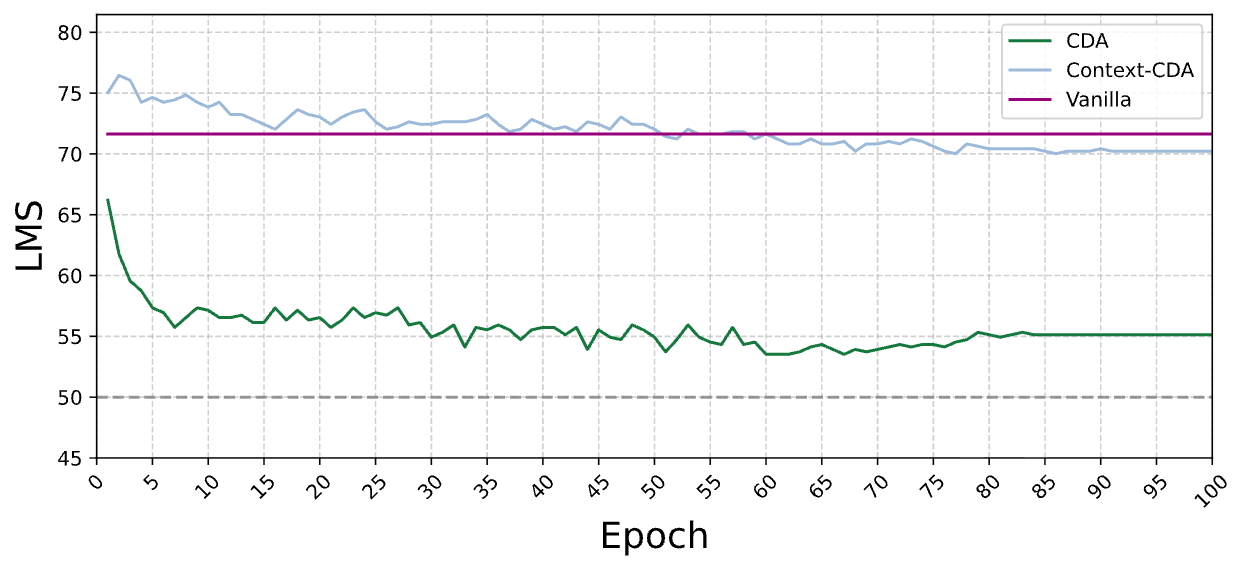}
    \caption{LMS score for T5. }
    \label{t5-lms}
\end{minipage}
\hfill
\begin{minipage}{0.48\linewidth}
    \centering
    \includegraphics[width=1\linewidth]{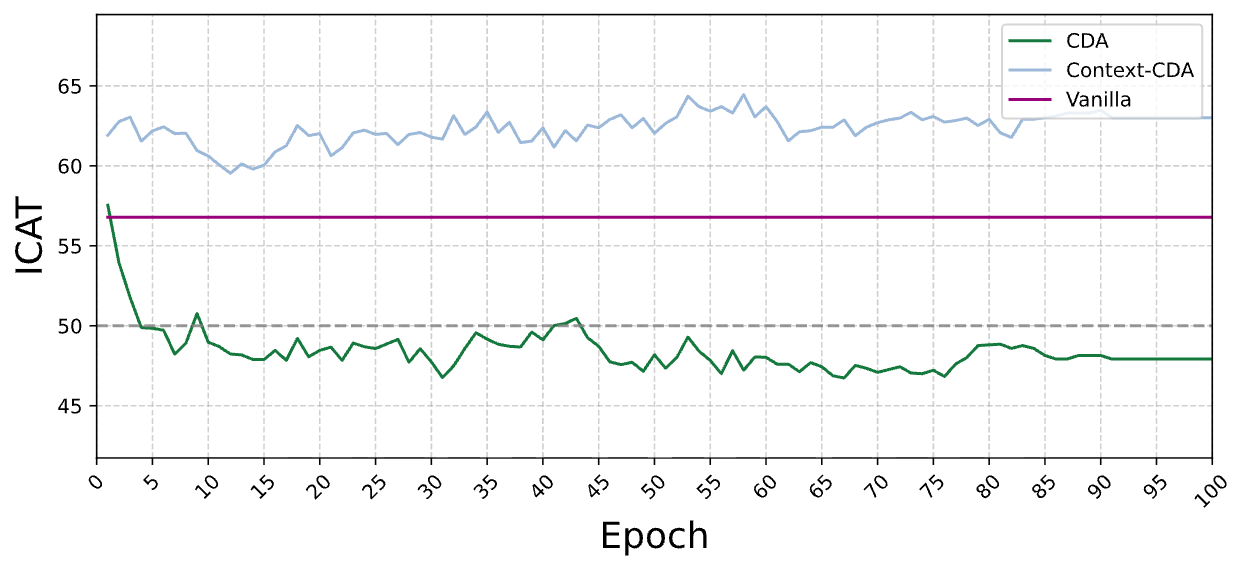}
    \caption{ICAT bias score for T5.}
    \label{t5-icat}
\end{minipage}
% \end{center}

\end{figure*}

\subsection{Uncertainty-Based Filtering}
\vspace{3ex}
With the generated counterfactuals, another challenge is that text generated by the larger LMs can be overly complex or noisy for the target smaller LMs to be debiased, hindering learning effectiveness. To further enhance the quality (e.g., mitigating hallucination) of the generated counterfactuals for the target LMs, we propose filtering out the generations that exhibit the greatest uncertainty by the target LMs. Uncertainty in the data has been shown to be an informative signal for algorithmic bias \cite{tahir2023fairness, singh2021fairness} and hallucination in large LMs \cite{farquhar2024detecting}. Specifically, we use a filtering process based on semantic entropy following the methodology in \cite{kuhn2023semantic}, which advances other uncertainty estimation approaches like likelihoods as it incorporates linguistic invariances created by shared meanings. For this process, we first sample multiple rephrased output sequences from a language model for each of our generated counterfactuals. These sequences are then clustered into semantic equivalence classes using a bidirectional entailment test following the methodology in \cite{kuhn2023semantic}, where two sequences are considered equivalent if they logically imply each other within context. Finally, the probabilities of sequences in each cluster are summed, and semantic entropy is calculated over these meaning-level probabilities to measure uncertainty over meanings rather than just surface forms. The semantic entropy (SE) for each counterfactual can be calculated as follows: 
\begin{equation}
    \label{eq:semantic_entropy}
    SE(\tilde{x}^c_i) \approx -|C|^{-1}\sum_{j=1}^{|C|} \log p(C_j \mid \tilde{x}^c_i),
\end{equation}
where $C$ denotes the number of samples generated by the larger LM. We calculate the semantic entropy for each generated counterfactual $\tilde{x}^c_i$ using the target LMs and filter out the top $k$-percent (e.g., 30\%) of sentences with the highest entropy. This ensures that overly complex or noisy sentences, which the target LMs may struggle to learn, are removed from the corpus, leaving only the most useful examples for debiasing during the fine-tuning.

\begin{figure*}[t]
% \begin{center}
    % \hfill
\centering
\begin{minipage}{0.48\linewidth}
    \centering
    \includegraphics[width=1\linewidth]{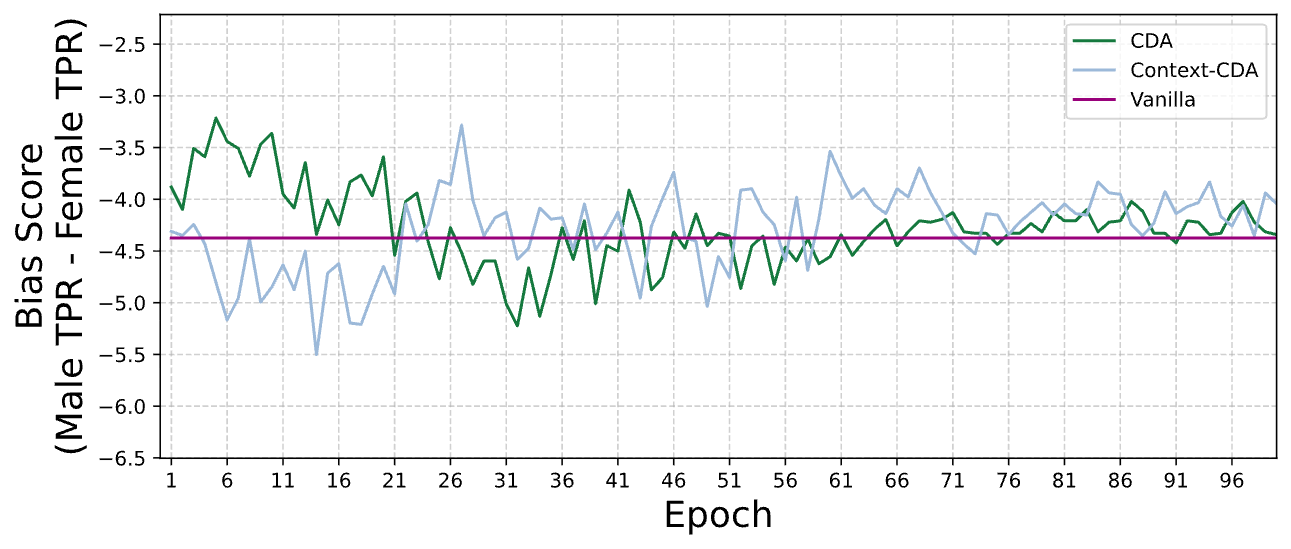}
    \caption{BiasBios score for BERT. 0 indicates no bias.}
    \label{bert-biasbios}
\end{minipage}
\hfill
\begin{minipage}{0.48\linewidth}
    \centering
    \includegraphics[width=1\linewidth]{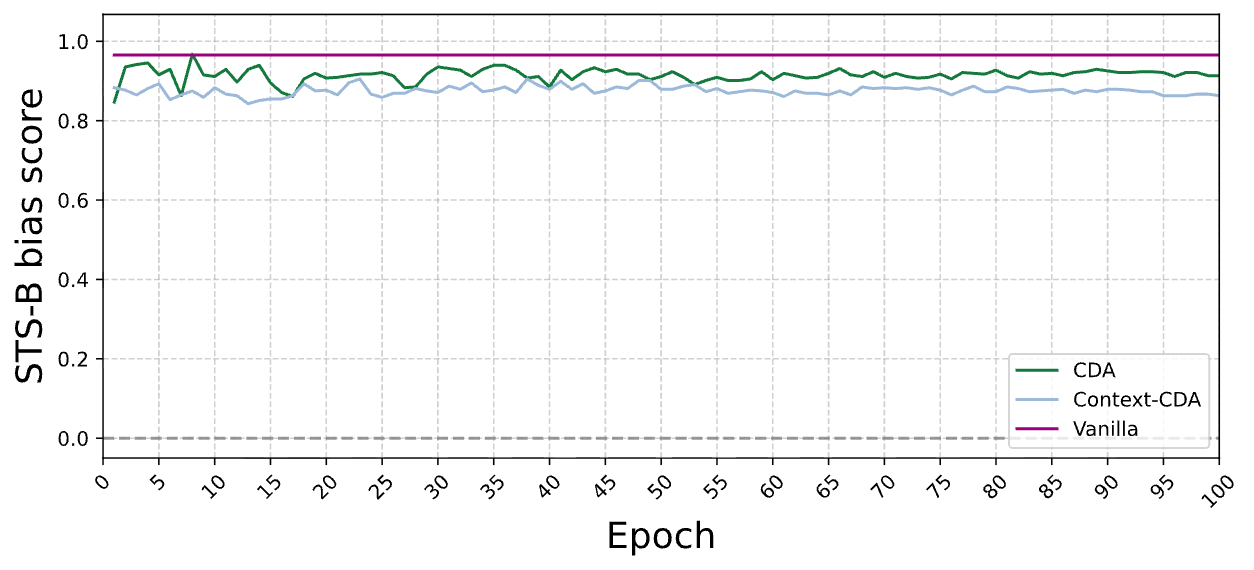}
    \caption{STS-B bias score for BERT. 0 indicates no bias.}
    \label{bert-stsb}
\end{minipage}
% \end{center}

% \begin{center}
\begin{minipage}{0.48\linewidth} % Keep the same width as before
    \centering
    \includegraphics[width=1\linewidth]{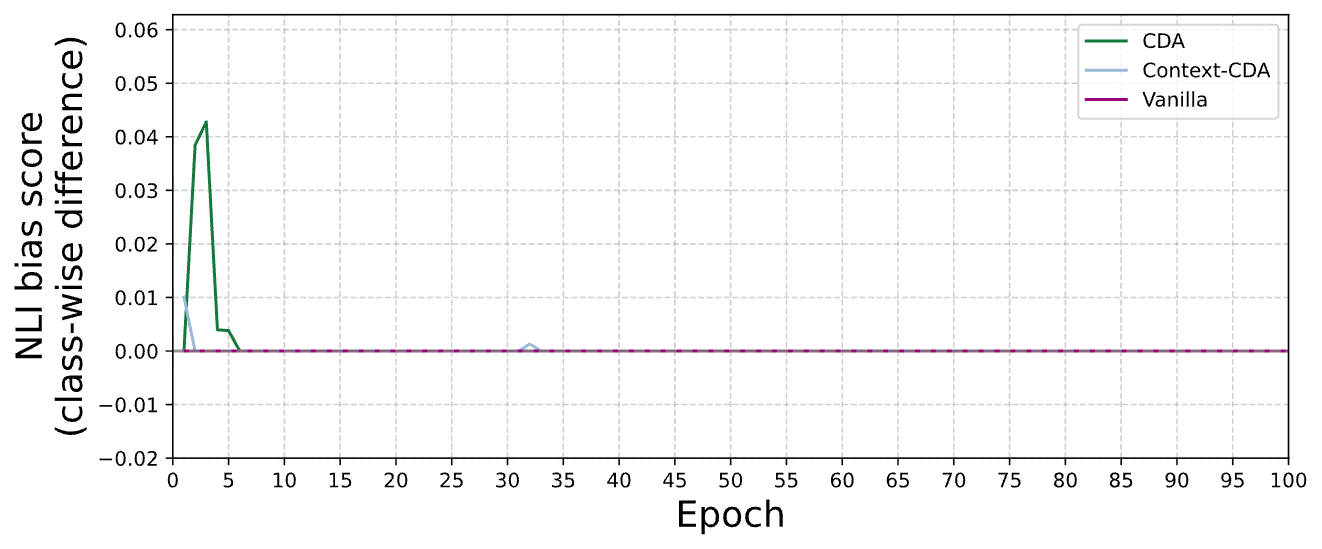}
    \caption{NLI-Bias score for BERT. 0 indicates no bias.}
    \label{bert-nlibias}
\end{minipage}
\hfill
\begin{minipage}{0.48\linewidth}
    \centering
    \includegraphics[width=1\linewidth]{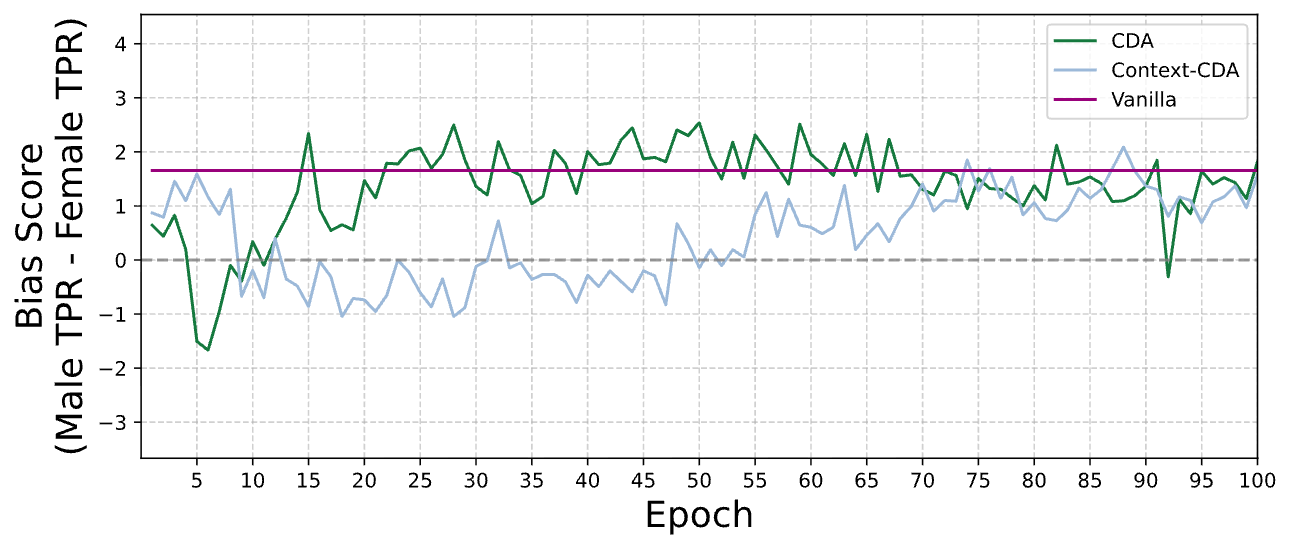}
    \caption{BiasBios score for GPT-2.}
    \label{gpt-biasbios}
\end{minipage}
% \end{center}

% \begin{center}
\hfill
\begin{minipage}{0.48\linewidth}
    \centering
    \includegraphics[width=1\linewidth]{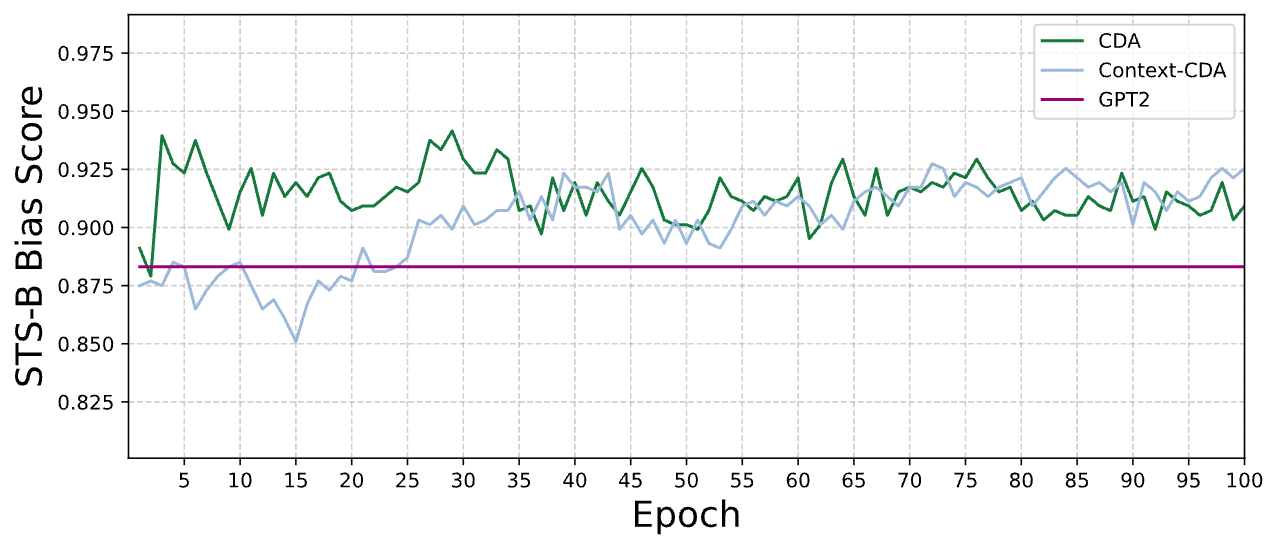}
    \caption{STS-B bias score for GPT-2.}
    \label{gpt-stsb}
\end{minipage}
\hfill
\begin{minipage}{0.48\linewidth}
    \centering
    \includegraphics[width=1\linewidth]{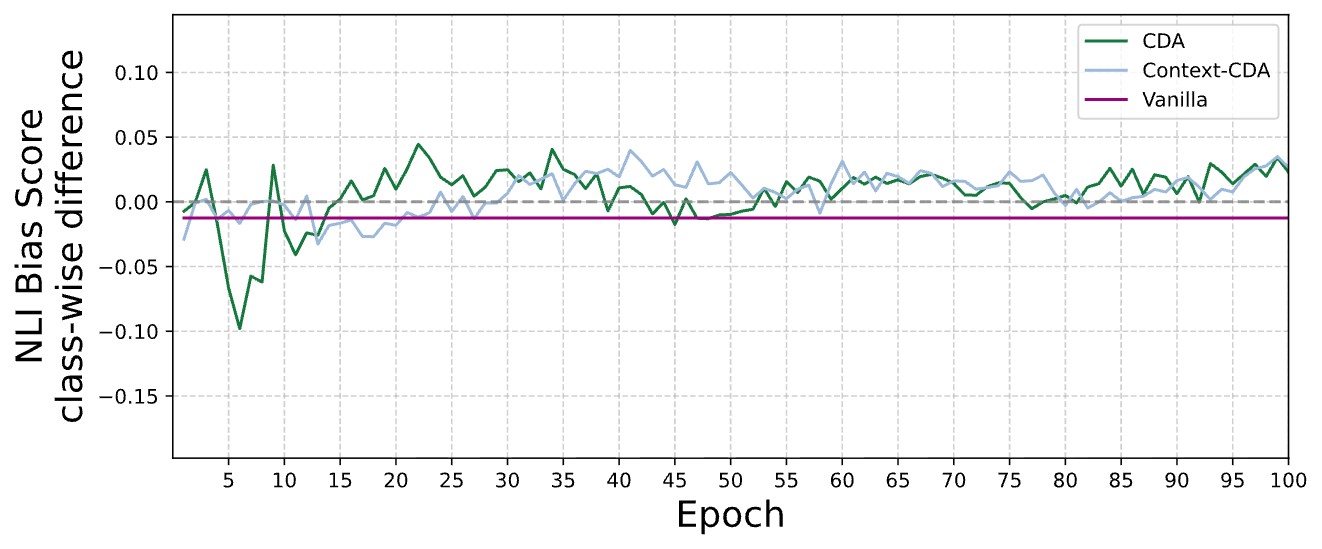}
    \caption{NLI-Bias score for GPT-2.}
    \label{gpt-nlibias}
\end{minipage}
% \end{center}

\end{figure*}

\subsection{Debiasing via Fine-tuning on Filtered \\Context-CDA}
\vspace{3ex}
We use the contextually rich counterfactual data samples obtained after augmentation and filtering to debias the target small LMs \cite{han2024chatgpt} like BERT and GPT-2. The key idea is to introduce alternative versions of the input data where specific attributes (e.g., gender) are modified without changing the underlying meaning via augmented context. During fine-tuning, this augmentation forces the model to learn representations that are invariant to these modifications, which helps mitigate biased correlations in the data. 

While our evaluation focuses on encoder models like BERT, this kind of representation debiasing is relevant to generative systems too. This is largely because BERT-like transformer encoders serve as foundational components in state-of-the-art encoder-decoder models (e.g., T5, BART, mBART) which are widely used for generation tasks such as machine translation and summarization. Moreover, contextualized encoder representations are commonly employed as frozen or fine-tuned backbones in modular and retrieval-augmented generation pipelines \cite{lewis2021retrievalaugmentedgenerationknowledgeintensivenlp}. As a result, debiasing these shared representations can directly benefit a broad range of downstream generative applications.

Moreover, to demonstrate generalizability of \textit{Context-CDA}, we evaluate across diverse architectures including encoder-only (BERT, DistilBERT), encoder-decoder (T5), and decoder-only (GPT-2, Llama-3.2-1B) models, as detailed in Section \ref{others}. In summary, our \textit{Context-CDA} method aims to balance debiasing and language modeling performance by generating augmented sentences that maintain linguistic coherence while reducing bias via fine-tuning. This ensures that the target model learns from a corpus that is both diverse and aligned with natural language patterns.

% and GPT-2 (decoder-only causal LM), as well as in extended experiments on DistilBERT (encoder-only, smaller, distilled model), T5 (encoder-decoder) and Llama-3-1B (causal language model), Context-CDA is architecture-agnostic and achieves effective debiasing across diverse model types, validating its applicability to both discriminative and generative settings.} 

\section{Experiments}
\label{others}
\vspace{3ex}
 We conduct the following experiments to validate the effectiveness of our proposed approach: (1) Evaluating the debiasing performance on intrinsic bias and language modeling capabilities, (2) Evaluating the debiasing performance on extrinsic bias, (3) Evaluating the debiasing performance on various downstream tasks, and (4) Analyzing next-token distribution to study predicted output token shifts after debiasing. All experiments are conducted across five diverse model architectures spanning encoder-only, encoder-decoder, and decoder-only designs to comprehensively validate \textit{Context-CDA}'s robustness and generalizability.

\begin{figure*}[t]
\centering
% \begin{center}
\begin{minipage}{0.48\linewidth}
    \centering
    \includegraphics[width=1\linewidth]{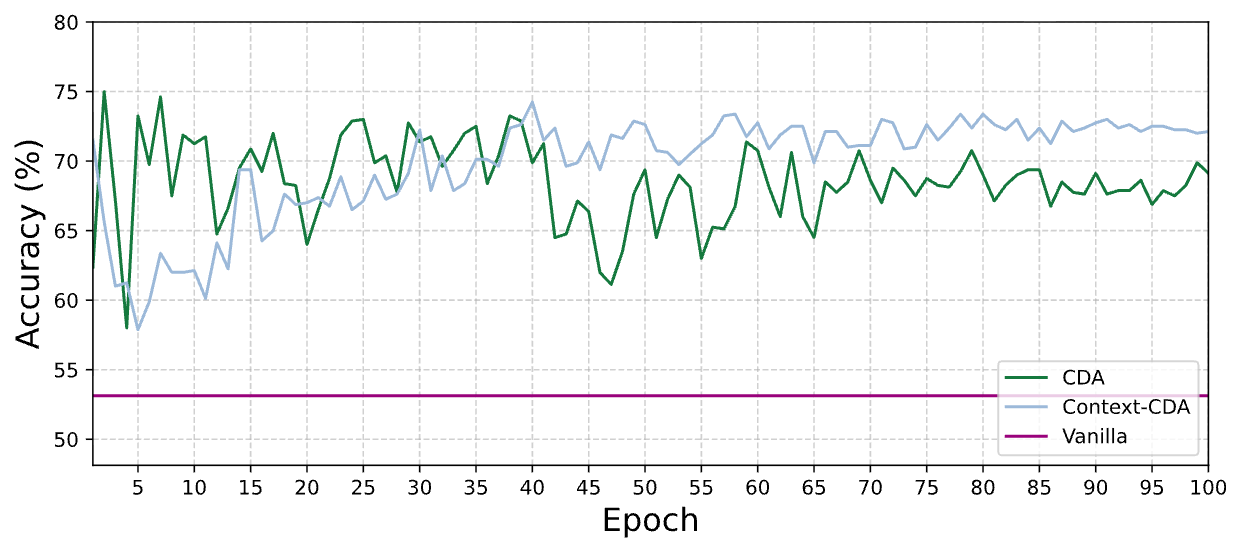}
    \caption{QNLI accuracy score ($\uparrow$) for BERT.}
    \label{bert-qnli}
\end{minipage}
\hfill
\begin{minipage}{0.48\linewidth}
    \centering
    \includegraphics[width=1\linewidth]{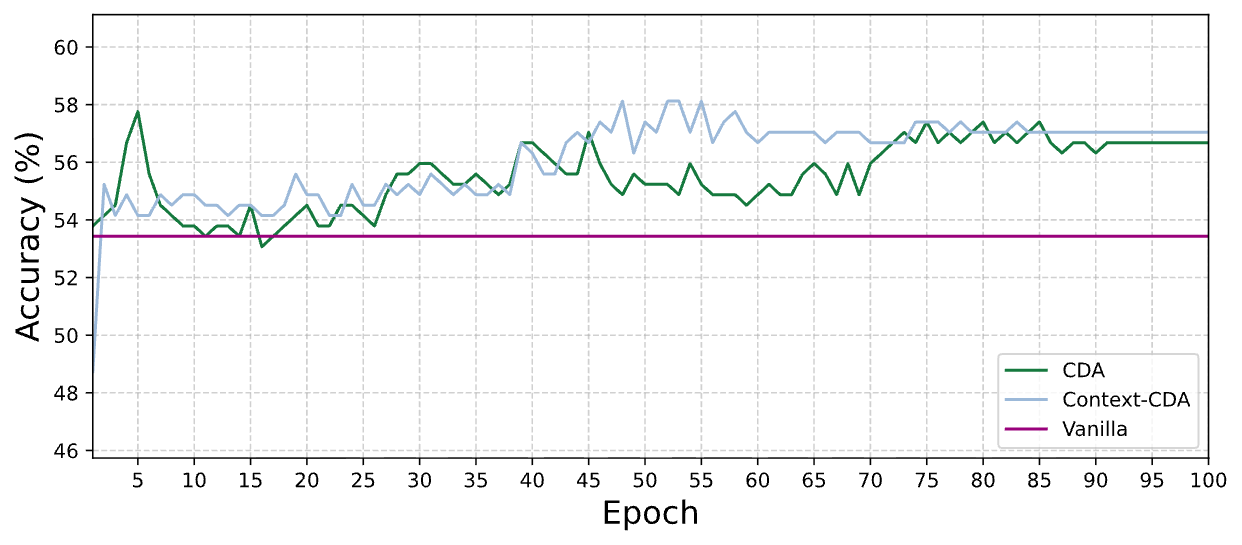}
    \caption{RTE scores ($\uparrow$) for BERT.}
    \label{bert-rte}
\end{minipage}
% \end{center}

% \begin{center}
\begin{minipage}{0.48\linewidth} % Keep the same width as before
    \centering
    \includegraphics[width=1\linewidth]{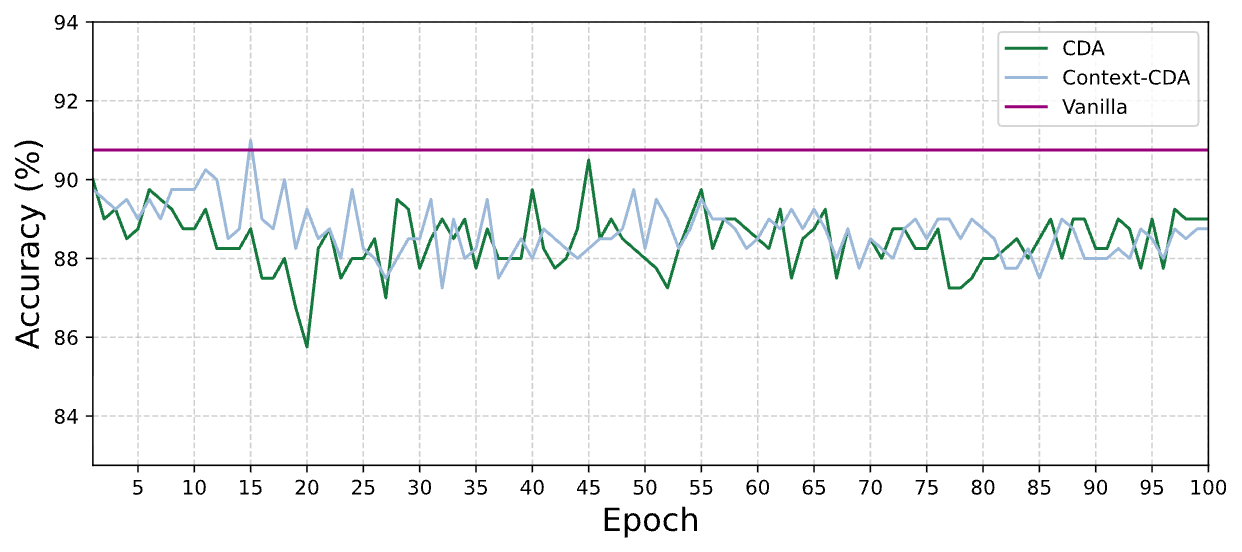}
    \caption{SST-2 accuracy score ($\uparrow$) for BERT.}
    \label{bert-sst2}
\end{minipage}
\hfill
\begin{minipage}{0.48\linewidth} % Keep the same width as before
    \centering
    \includegraphics[width=1\linewidth]{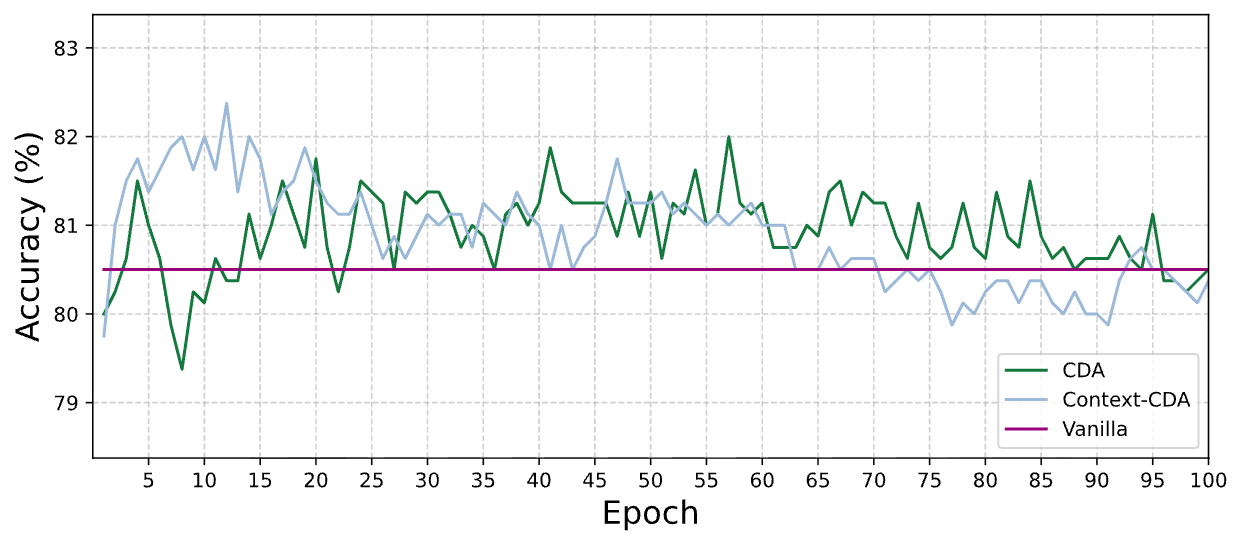}
    \caption{QNLI accuracy score for GPT-2.}
    \label{gpt-qnli}
\end{minipage}
% \end{center}

% \begin{center}
\begin{minipage}{0.48\linewidth} % Keep the same width as before
    \centering
    \includegraphics[width=1\linewidth]{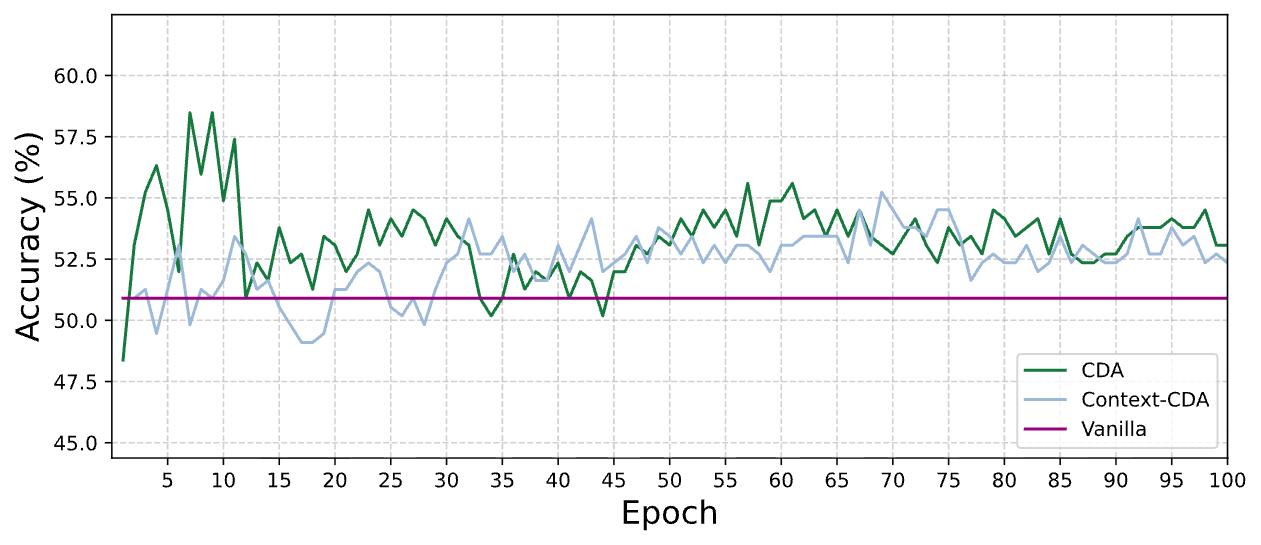}
    \caption{RTE accuracy score for GPT-2.}
    \label{gpt-rte}
\end{minipage}
\hfill
\begin{minipage}{0.48\linewidth} % Keep the same width as before
    \centering
    \includegraphics[width=1\linewidth]{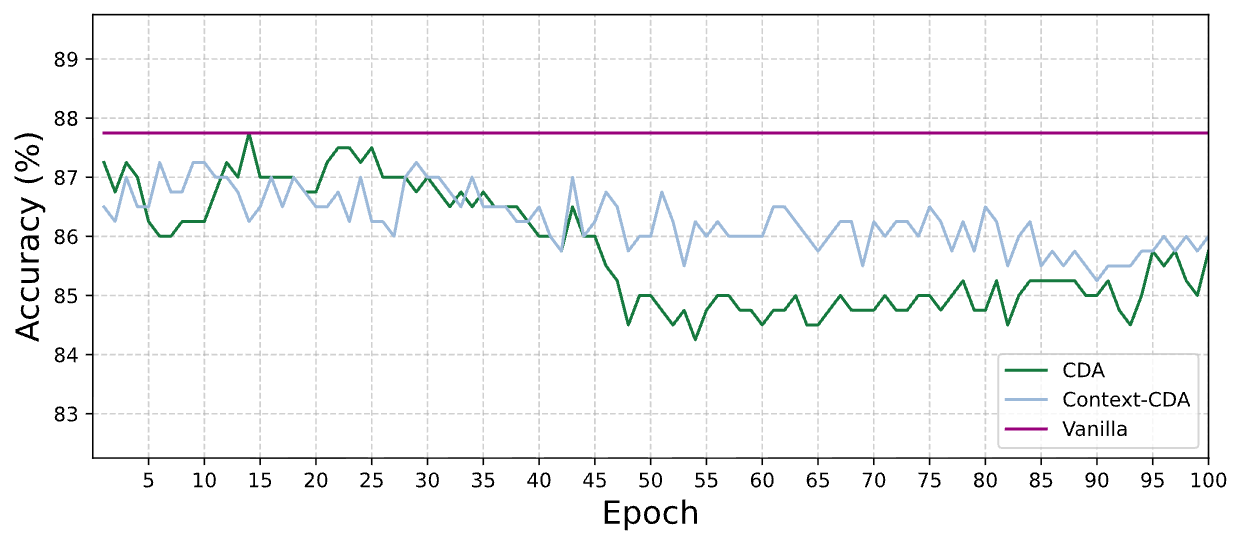}
    \caption{SST-2 accuracy score for GPT-2.}
    \label{gpt-sst2}
\end{minipage}
% \end{center}

\end{figure*}

\subsection{Experimental Setup}
\vspace{3ex}
\textbf{Datasets: } To implement gender bias mitigation, we use the news-commentary dataset \cite{website1} as our debiasing corpus following \cite{kaneko2022debiasing} with the gender words list from \cite{zhao2018learning}. For \textbf{intrinsic bias}, we use gender samples from two benchmark datasets: (1) StereoSet \cite{nadeem2020stereoset} tests whether models can complete sentences without reinforcing harmful stereotypes, while still maintaining language fluency. (2) CrowS-Pairs \cite{nangia2020crows} tests social biases in LMs by presenting sentence pairs, where one reflects a stereotype and the other does not. For \textbf{extrinsic bias}, we use datasets (1) BiasBios (Bias in Bios) \cite{de2019bias}, (2) STS-B (Semantic Textual Similarity Benchmark) \cite{cer2017semeval} and (3) NLI-Bias (Natural Language Inference-Bias) \cite{de2019bias}. For evaluating model performances on downstream tasks like sentiment analysis, entailment, and question-answering, we use the following datasets from the GLUE (General Language Understanding Evaluation) benchmark \cite{wang2018glue}, (1) SST-2 (Stanford Sentiment Treebank-2) \cite{socher2013recursive}, (2) RTE (Recognizing Textual Entailment) \cite{dagan2005pascal}, and (3) QNLI (Question-answering Natural Language Inference) \cite{wang2018glue} derived from the Stanford Question Answering Dataset (SQuAD).

% \vspace{0.3\baselineskip}
\par\vspace{0.3em}
% \noindent\textbf{Base LMs: } 
%Following \cite{}, 
% We use two classic LMs, BERT \cite{devlin2019bert} and GPT-2 \cite{radford2019language}, for intrinsic and extrinsic bias evaluation, and downstream tasks. 
\noindent\textbf{Base LMs: } Our evaluation covers five diverse LMs spanning different architectures. This includes two encoder-only models (BERT \cite{devlin2019bert} and DistilBERT \cite{sanh2020distilbertdistilledversionbert}), one encoder-decoder model (T5 \cite{raffel2023exploringlimitstransferlearning}), and two causal decoder-only models (GPT-2 \cite{radford2019language} and Llama-3.2-1B \cite{huggingfaceLlama3}) for a comprehensive evaluation. We evaluate BERT and GPT-2 on all intrinsic bias, extrinsic bias, and downstream task performance metrics. DistilBERT, T5, and Llama-3.2-1B are evaluated solely on intrinsic bias metrics. This selection enables us to validate that \textit{Context-CDA} is model-agnostic and effective across encoder-only, encoder-decoder, and decoder-only architectures.

% \vspace{0.3\baselineskip}
\par\vspace{0.3em}
\noindent\textbf{Baselines:} We compare the Vanilla model, which is the pre-trained LM without debiasing, and LMs debiased using traditional CDA as the baseline for all models including BERT, GPT-2, DistilBERT, T5 and Llama-3.2-1B. Additionally, we consider several baselines for BERT and GPT-2, respectively.
For BERT, we use the following methods as the baseline: (1) MABEL \cite{he2022mabel}, an intermediate pre-training approach for mitigating gender bias in contextualized representation; (2) INLP \cite{ravfogel2020null}, mitigating gender bias in word embeddings by removing information from neural representations; (3) SelfDebias \cite{schick2021selfdiagnosisselfdebiasingproposalreducing}, a decoding algorithm that, given only a textual description of the undesired behavior, reduces the probability of an LM producing problematic text; and (4) SENT-DEBIAS \cite{liang-etal-2020-towards}, which reduces gender bias in sentence-level representations. For GPT-2, we use the following methods as baselines: (1) SelfDebias \cite{schick2021selfdiagnosisselfdebiasingproposalreducing}, a decoding algorithm that, given only a textual description of the undesired behavior, reduces the probability of an LM producing problematic text; (2) SENT-DEBIAS \cite{liang-etal-2020-towards}, which reduces gender bias in sentence-level representations, and (3) wiki-debiased \cite{xie2023empiricalanalysisparameterefficientmethods}, a baseline with GPT-2 as the base LM that uses parameter-efficient methods to fine-tune GPT-2 using WikiText2 \cite{merity2016pointersentinelmixturemodels}.

\begin{table*}
  \centering
  \small
  \arrayrulecolor{black}
  \renewcommand{\arraystretch}{1.2}
  \setlength{\arrayrulewidth}{0.3mm}
  \setlength{\tabcolsep}{5pt}

  % \begin{minipage}{.5\linewidth}
  % \centering
  \begin{tabular}{|c|c|c|c|c|c|c|c|c|}
    \hline

    \rowcolor{gray!30}
    & \multicolumn{4}{c|}{\textbf{BERT}} & \multicolumn{4}{c|}{\textbf{GPT-2}} \\
    \hline
    \rowcolor{gray!30}
    \textbf{Debiasing Technique} & \textbf{SS} &  \textbf{LMS} ($\uparrow$) & \textbf{ICAT} ($\uparrow$)& \textbf{CS} & \textbf{SS} &  \textbf{LMS} ($\uparrow$)& \textbf{ICAT} ($\uparrow$)& \textbf{CS} \\
    \hline
    MABEL & 47.28 & 51.65 & 48.84 &  \underline{52.29} & - & - & - & -             \\
    INLP & \underline{49.16} & 50.25 & 49.41  & 55.73  & - & - & - & - \\
    wiki-debiased & - & - & - & - & 60.40 & \underline{91.01} & 72.08 & 56.49 \\
    SelfDebias &  59.34 & \underline{84.20} & \underline{68.47} & \underline{52.29}   & \underline{56.05}  & 87.43 & \underline{73.18} & 56.11 \\
    SENT-DEBIAS & 59.37 & 84.09 & 68.33 & \underline{52.29}  & 60.84 & 89.07 & 69.76 & 56.11 \\
    Vanilla &  59.95 & 79.87 & 63.96  & 58.01  & 63.17 & 77.46 & 57.04 & 51.52 \\
    CDA &  58.55 & 68.41 &  56.71 & 54.96   & 57.34 & 69.61  & 59.39 & \underline{50.01} \\
    \textit{Context-CDA} & 57.75 & 78.67 &  66.48 & 53.43  & 56.13 & 75.25 & 66.01 & 50.76 \\
    \hline
  \end{tabular}
  \caption{\label{MLM-intrinsic-bias}
    Intrinsic bias evaluation for BERT and GPT-2 comparing baselines. Underlined values indicate the best performance.}
  % \end{minipage}
\label{bert_gpt_baseline}  
\end{table*}

% \vspace{0.3\baselineskip}
\par\vspace{0.3em}
\noindent\textbf{Metrics: } 
%Following \cite{}, 
To evaluate debiasing performance, we distinguish between intrinsic bias and extrinsic bias. Intrinsic bias refers to the stereotypical associations encoded directly within a language model’s representations or probabilities, independent of downstream tasks. It is commonly measured through benchmarks such as StereoSet and CrowS-Pairs, which assess whether models favor stereotypical over anti-stereotypical continuations or sentence pairs. Extrinsic bias captures disparities that arise when models are applied to downstream tasks. It reflects whether representational biases affect task performance, for instance, differences in prediction accuracy or true positive rates across demographic groups. We measure extrinsic bias using datasets such as BiasBios, STS-B, and NLI-Bias. 

\par\vspace{0.3em}
For intrinsic bias, we report four standard metrics from the StereoSet \cite{nadeem2020stereoset} and CrowS-Pairs \cite{nangia2020crows} benchmarks: (1) Stereotype Score (SS) calculates the percentage of examples in which a model prefers a stereotypical association over an anti-stereotypical association (a score of 50 indicates no bias and a score above and below 50 indicates preference for stereotypical or anti-stereotypical associations); (2) Language Modeling Score (LMS) calculates the percentage of instances in which an LM prefers meaningful over meaningless association (a higher score indicates better language performance); (3) Idealized CAT Score (ICAT) describes the comprehensive performance of the model by combining SS and LMS into one score (a higher score indicates better performance on debiasing while maintaining language fluency); and (4) CrowS-Pairs Score (CS) calculates the percentage of instances where the model assigns a higher probability to the stereotypical sentence over the anti-stereotypical one (a score closer to 50 indicates less bias, while scores above or below 50 suggest a preference toward biased associations).

\par\vspace{0.3em}
For extrinsic bias, we report the following three metrics: (1) NLI-Bias measures the class-wise difference between male and female samples in the NLI-Bias dataset, with scores closer to 0 indicating less bias; (2) BiasBios  measures the difference in true positive rate between male and female samples in BiasBios dataset; and (3) STS-B flips the gendered words in the sentence pairs and then calculates semantic similarity in original and gender-flipped sentence pairs to measure gender bias. The bias score is calculated as the difference in prediction accuracy between male and female groups, with scores closer to 0 indicating less bias. For performance on downstream tasks, we use accuracy as a metric for datasets (1) QNLI, (2) RTE, and (3) SST-2.

\vspace{0.3\baselineskip}

\par\vspace{0.3em}
We set the uncertainty filtering threshold $k$ to 30\% as it gives the best performance based on our debiasing corpus as shown in the ablation study in Appendix \ref{ablation_study}. For each language model, including BERT, GPT-2, DistilBERT, T5, and Llama-3.2-1B, we perform debiasing at each epoch to evaluate the intrinsic bias and language modeling performance using the StereoSet and CrowS-Pairs benchmarks. Since our method fine-tunes models, we compare the vanilla, CDA, and \textit{Context-CDA} models at each epoch of debiasing. For BERT and GPT-2, we additionally fine-tune models at each epoch to measure extrinsic bias and downstream task performance across datasets. This ensures a comprehensive assessment of debiasing effectiveness across both representation-level and task-level biases. This multi-faceted evaluation enables us to assess not only whether bias is reduced but also whether language modeling capability and downstream task performance are preserved.

\subsection{Post Debiasing Performance on Intrinsic Bias Scores and Language Modeling}
\vspace{3ex}

\subsubsection{Multi-Model Evaluation: Demonstrating Robustness and Generalizability}

\vspace{3ex}
% \par\vspace{0.3em}

To validate the robustness and model-agnostic nature of \textit{Context-CDA}, we evaluate intrinsic bias on all five LMs.

\textbf{Intrinsic Bias.} To measure intrinsic bias, we conduct evaluations using the StereoSet and CrowS-Pairs benchmark at every debiasing epoch. We present detailed results of CDA and \textit{Context-CDA} across models: BERT (Figs.~\ref{bert-ss} and \ref{bert-cs}), DistilBERT (Figs.~\ref{distilbert-ss} and \ref{distilbert-cs}), GPT-2 (Figs.~\ref{gpt-ss} and \ref{gpt-cs}), Llama-3.2-1B (Figs.~\ref{llama-ss} and \ref{llama-cs}), and T5 (Figs.~\ref{t5-ss} and \ref{t5-cs}). Here, scores closer to 50 indicate less bias with a score of 50 indicating no bias. We observe that the vanilla BERT and DistilBERT generally start with the highest bias scores before debiasing begins. As debiasing progresses across epochs, both encoder models show consistent debiasing patterns and the SS and CS scores gradually approaches 50, indicating a reduction in bias in the models. We also observe that towards the end of training epochs, the bias scores stabilize and we conclude that further fine-tuning is not required. Towards epochs 75-85, \textit{Context-CDA} starts outperforming CDA and achieves a stable bias score better than CDA. This consistency validates \textit{Context-CDA} for encoder-only architectures. In GPT-2, CS bias score for \textit{Context-CDA} and traditional CDA reaches around 50 (no bias) early on and oscillates about the same mean bias score as fine-tuning progresses. In Llama-3.2-1B, epochs 0 to 25 show a sustained decline in SS score reaching a plateau that indicates convergence without overfitting. The stable plateau throughout training demonstrates that further fine-tuning does not degrade or improve the debiasing performance. Both generative language models achieve comparable or superior results to their non-augmented baselines, supporting our claim that \textit{Context-CDA} is effective for generative systems as well. T5, as an encoder-decoder model, shows particularly strong performance, achieving a balanced SS score and a significant improvement in CS score. Thus, compared to traditional CDA, \textit{Context-CDA} shows consistent improvement across all intrinsic metrics.

\par\vspace{0.3em}
\textbf{Language Modeling Ability.} To measure language modeling performance, we evaluate the LMS and ICAT scores at every debiasing epoch. We present detailed results of CDA and \textit{Context-CDA} across models: BERT (Figs.~\ref{bert-lms} and \ref{bert-icat}), DistilBERT (Figs.~\ref{distilbert-lms} and \ref{distilbert-icat}), GPT-2 (Figs.~\ref{gpt-lms} and \ref{gpt-icat}), Llama-3.2-1B (Figs.~\ref{llama-lms} and \ref{llama-icat}), and T5 (Figs.~\ref{t5-lms} and \ref{t5-icat}). Notably, across all models, \textit{Context-CDA} consistently improves both LMS and ICAT scores compared to CDA. \textit{Context-CDA} consistently outperforms CDA and achieves LMS scores as well as Vanilla models, while outperforming both Vanilla and CDA baselines in ICAT scores. This is a strong indicator of \textit{Context-CDA's} superior language modeling performance compared to CDA, reinforcing that \textit{Context-CDA} achieves better linguistic understanding and better preserves language modeling capability while reducing bias. We attribute this improvement to the greater diversity and naturalness of the \textit{Context-CDA} corpus compared to traditional CDA, which enables models to retain stronger linguistic representations after debiasing. Overall, \textit{Context-CDA} achieves effective bias mitigation while simultaneously improving language modeling performance.

% \subsubsection{Convergence Patterns}

% \vspace{3ex}

% \textcolor{red}{Across all five models, debiasing performance stabilizes at epochs 75-85. This convergence is significant: it demonstrates that  (1) the debiasing process reaches a stable plateau, indicating the model has learned the debiased representations without overfitting; (2) further training beyond this point provides diminishing returns, suggesting efficient training dynamics; and (3) the stability is not an artifact of a single model, as the same pattern holds across BERT, DistilBERT, GPT-2, Llama-3-1B, and T5, strengthening the claim that Context-CDA achieves robust, generalizable debiasing.}

\subsubsection{Convergence Patterns}

\vspace{3ex}

The per-epoch analysis reveals important training dynamics: (1) \textbf{Convergence Timing:} Across all five models, debiasing performance stabilizes at epochs 75-85, indicating the method reaches optimal performance earlier in training. (2) \textbf{Stability Without Overfitting:} The plateau in bias scores and sustained or improved scores demonstrate that models do not overfit to the debiasing corpus; instead, they learn stable, debiased representations. (3) \textbf{Model-Agnostic Pattern:} The identical convergence behavior across BERT, DistilBERT, GPT-2, Llama-3.2-1B, and T5 validates that this stability is not an artifact of a single architecture but a general property of \textit{Context-CDA}. (4) \textbf{Training Efficiency:} The early stabilization suggests that practitioners can reduce training epochs, making the method computationally efficient. These insights strengthen confidence in the method's robustness and generalizability.

\subsubsection{Comparison with Prior Debiasing Methods}

\vspace{3ex}

\par\vspace{0.3em}
We also compare the bias scores of fine-tuned BERT and GPT-2 against other debiasing baselines in Table \ref{bert_gpt_baseline} for an overall comparison. While methods such as SelfDebias and SENT-DEBIAS achieve higher scores on isolated metrics (e.g., ICAT), they often rely on inference-time interventions or post-hoc representation manipulation, which: (1) are model-specific or task-specific, limiting generalizability; (2) do not demonstrate consistent performance across both intrinsic and extrinsic metrics; and (3) lack validation on diverse model architectures. In contrast, our extended evaluation demonstrates that \textit{Context-CDA} achieves balanced, consistent performance across five distinct architectures, maintaining stable debiasing while preserving language modeling capability across both encoder-only and generative models. Wiki-debiased, in contrast, leverages curated Wikipedia data, making it domain-limited, whereas \textit{Context-CDA} can be applied to arbitrary corpora. Consequently, \textit{Context-CDA} is model-agnostic, requires no architecture modifications, and integrates seamlessly into the fine-tuning pipeline, making it a practical and effective debiasing solution across model architectures. The consistency of results across five model types provides strong empirical evidence of robustness, addressing concerns about limited generalization from single-model evaluation. It strikes a strong balance by offering the best trade-off between bias mitigation and language modeling performance, outperforming standard CDA across all intrinsic bias metrics and achieving comparable or superior results to other baselines, all without sacrificing fluency.

\begin{figure*}[t]
\begin{center}
%\framebox[4.0in]{$\;$}
% \fbox{\rule[-.5cm]{0cm}{4cm} \rule[-.5cm]{4cm}{0cm}}
\includegraphics[width=0.9\textwidth]{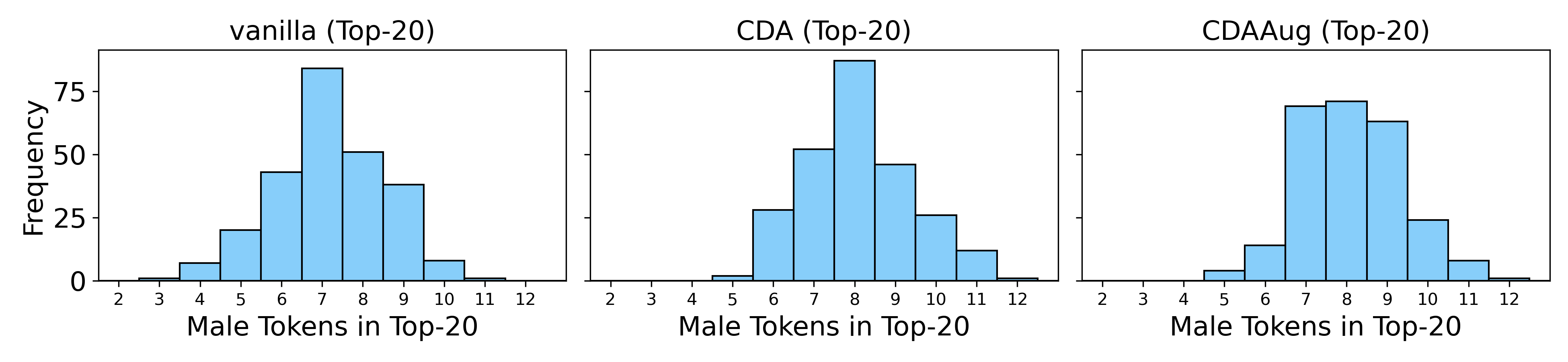} % Adjust width & height as needed
\end{center}
\caption{Top-20 male token distributions for GPT2.}
\label{token_dist_male}
% \end{figure*}

% \begin{figure*}[t]
\begin{center}
%\framebox[4.0in]{$\;$}
% \fbox{\rule[-.5cm]{0cm}{4cm} \rule[-.5cm]{4cm}{0cm}}
\includegraphics[width=0.9\textwidth]{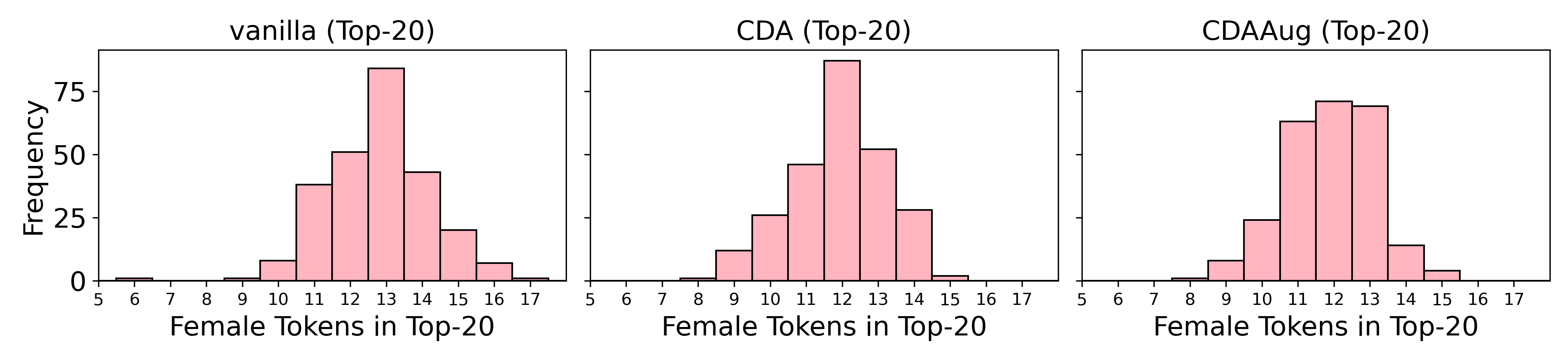} % Adjust width & height as needed
\end{center}
\caption{Top-20 female token distributions for GPT2.}
\label{token_dist_female}
\end{figure*}

\subsection{Post Debiasing Performance on Extrinsic Bias and Downstream Tasks}
\vspace{3ex}
%\z{Why you only compare CDA and Context-CDA, where are other baselines?}

For evaluating performance on extrinsic bias metrics, we further fine-tune our models on STS-B, NLI-Bias and BiasBios tasks and then evaluate the extrinsic bias scores on the fine-tuned models. Here, scores closer to 0 indicate low bias. For BERT, we find that for STS-B (Fig.~\ref{bert-stsb}), \textit{Context-CDA} performs better than CDA whereas for NLI-Bias and BiasBios (Fig.~\ref{bert-nlibias} and Fig.~\ref{bert-biasbios} respectively) \textit{Context-CDA} performs as good as CDA indicating that \textit{Context-CDA} is robust for even extrinsic bias evaluation metrics. For GPT-2, \textit{Context-CDA} performs as well as CDA for BiasBios, STS-B, and NLI-Bias (Figs.~\ref{gpt-biasbios}-\ref{gpt-nlibias}), further validating its effectiveness in reducing extrinsic bias across model architectures. For evaluating performance on various downstream language understanding tasks, we further fine-tune our debiased models on QNLI, RTE, and SST-2 and then evaluate the fine-tuned models for accuracy. Higher score indicates higher accuracy. For QNLI (Fig.~\ref{bert-qnli}), \textit{Context-CDA} performs better than CDA whereas for RTE and SST-2 (Fig.~\ref{bert-rte} and Fig.~\ref{bert-sst2}), \textit{Context-CDA} performs slightly better or as good as CDA, indicating that models fine-tuned with \textit{Context-CDA} can perform better than CDA even in various downstream tasks.

\subsection{Next-Token Distribution}
\vspace{3ex}
To gain deeper insights into the debiasing performance, we compare the vanilla LM, CDA, and \textit{Context-CDA} by examining next-token distributions in gender-related contexts. We focus on the GPT-2 model with the Stereotype Score (SS) closest to 50 after CDA debiasing, allowing us to investigate how the debiasing process affects token distribution and explore its impact on model predictions. 
We use sentences from StereoSet that specifically evaluate gender bias. Each sentence is split into two parts: the portion before the BLANK, referred to as the context, and the portion after it. The context is fed into GPT-2, and we compute the logits for the next token, extracting scores for tokens in a predefined male–female mapping set  (approximately 200 words) \cite{zhao2018learning}. After applying softmax, we obtain probabilities for each word and identify the top-20 tokens. We then count how many tokens are male-related and how many are female-related.

\par\vspace{0.3em}
Figures \ref{token_dist_male} and \ref{token_dist_female} illustrate the frequency of male and female tokens, respectively, among the top-20 predictions. For male tokens, the vanilla LM peaks at 7 male-related tokens, while CDA and \textit{Context-CDA} shift the peak to 8. Importantly, \textit{Context-CDA} yields a more balanced distribution across contexts, reducing skew toward male-associated tokens. Similarly, for female tokens, the vanilla LM peaks at 13 female-related tokens, while CDA and \textit{Context-CDA} shift the peak toward 12. Again, \textit{Context-CDA} produces a more even distribution, avoiding over-concentration on specific female-related tokens. 
Together, these results demonstrate that both CDA and \textit{Context-CDA} adjust the token distribution away from the stronger bias seen in the Vanilla model. However, \textit{Context-CDA} is more effective at spreading token probabilities evenly across male and female categories. This not only mitigates over-reliance on gender-specific terms but also promotes linguistic diversity, thereby improving robustness in predictions and reflecting a more balanced language modeling ability.

\section{Limitations}
\vspace{3ex}
While our method \textit{Context-CDA} demonstrates improvements over traditional CDA in mitigating gender bias without compromising language modeling performance, several limitations remain. First, using large LMs for generating context-rich augmentations may introduce computational and environmental costs, which may hinder scalability and accessibility in resource-constrained settings. Second, although semantic entropy filtering improves corpus quality by excluding uncertain examples, it may also remove valuable complex counterfactuals; future work could explore adaptive or multi-criteria filtering strategies that balance quality, diversity, and training stability. Third, our study focuses primarily on binary gender counterfactuals. Extending this framework to non-binary and intersectional identities, as well as other sensitive attributes (e.g., race, religion, or profession), is an important next step. Fourth, because large LMs and target smaller models may differ in distributional characteristics, alignment mechanisms such as domain-adaptive filtering, grounding mechanisms such as fact-checking modules, and human-in-the-loop validation could further enhance contextual reliability. Finally, expanding this framework to include domain-specific bias detection or multilingual or multimodal extensions would enhance robustness and fairness across broader applications.

\section{Conclusion}
\vspace{3ex}
This work presents an effective gender debiasing method that maintains competitive performance in downstream tasks while reducing bias in LMs. Building on classic CDA, which effectively mitigates bias but often weakens language modeling capabilities, our proposed method, \textit{Context-CDA}, enhances the debiasing corpus by leveraging large LMs to generate enriched context. This augmentation minimizes the discrepancy between the debiasing corpus and the original pre-training data, ensuring better alignment and also preserving linguistic fluency. Furthermore, we incorporate semantic entropy filtering to remove uncertain content, improving the overall quality of the generated corpus. Comprehensive evaluation across five diverse model architectures  demonstrates that \textit{Context-CDA} is truly model-agnostic, achieving robust and consistent debiasing performance across both discriminative and generative systems. Our method not only mitigates bias effectively but also enhances language modeling performance. As LMs continue to evolve, integrating more sophisticated debiasing techniques will be crucial for building more equitable and more robust AI systems.

\bibliographystyle{abbrv}
\bibliography{aaai25.bib}   % sigproc.bib is the name of the Bibliography in this case
% ACM needs 'a single self-contained file'!
% SIGKDD: balancing columns messes up the footers: Sunita Sarawagi, Jan 2000.
% \balancecolumns
\appendix

\section{Ablation Study on Uncertainty Threshold} 
\label{ablation_study}
\vspace{3ex}

To validate the robustness of our proposed \textit{Context-CDA} approach, we conduct an ablation study analyzing the effect of varying the semantic entropy filtering threshold on debiasing and language modeling performance. In our main experiments, we used a default threshold of 30\%, removing the top 30\% of counterfactuals with the highest semantic entropy. This study evaluates how different thresholds (20\% and 40\%) influence the effectiveness of the model.

% \vspace{0.3\baselineskip}
\par
\begin{figure*}[t]
\centering
% \begin{center}
\begin{minipage}{0.48\linewidth}
    \centering
    \includegraphics[width=1\linewidth, height=3cm,
  keepaspectratio=false]{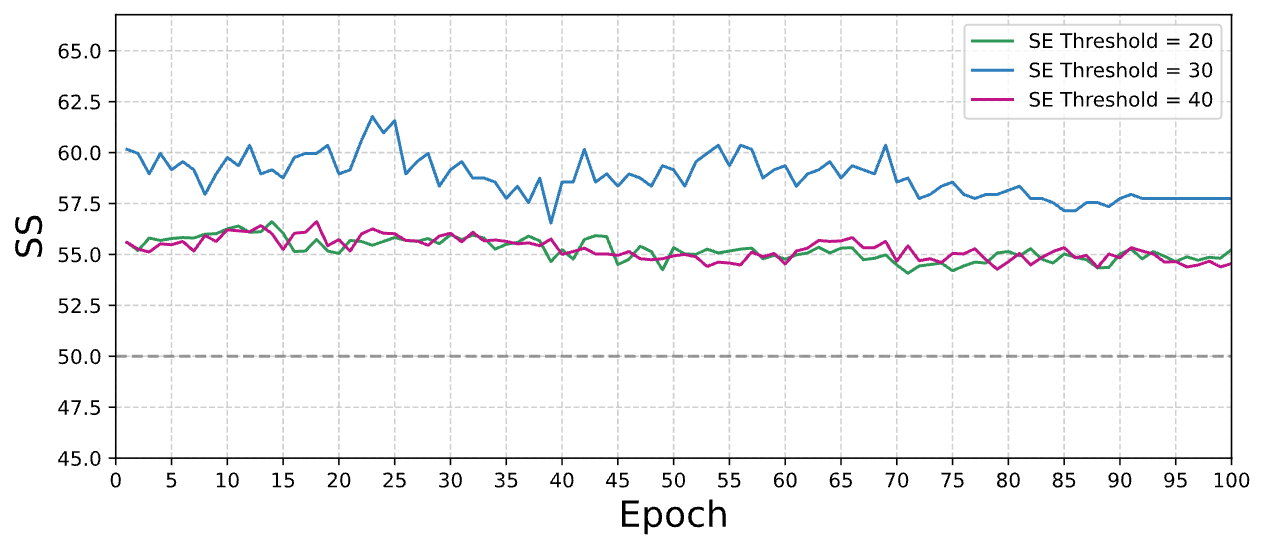}
    \caption{BERT SS Score for SE thresholds 20, 30 and 40.}
    \label{bert-203040-ss}
\end{minipage}
\hfill
\begin{minipage}{0.48\linewidth} % Keep the same width as before
    \centering
    \includegraphics[width=1\linewidth, height=3cm,
  keepaspectratio=false]{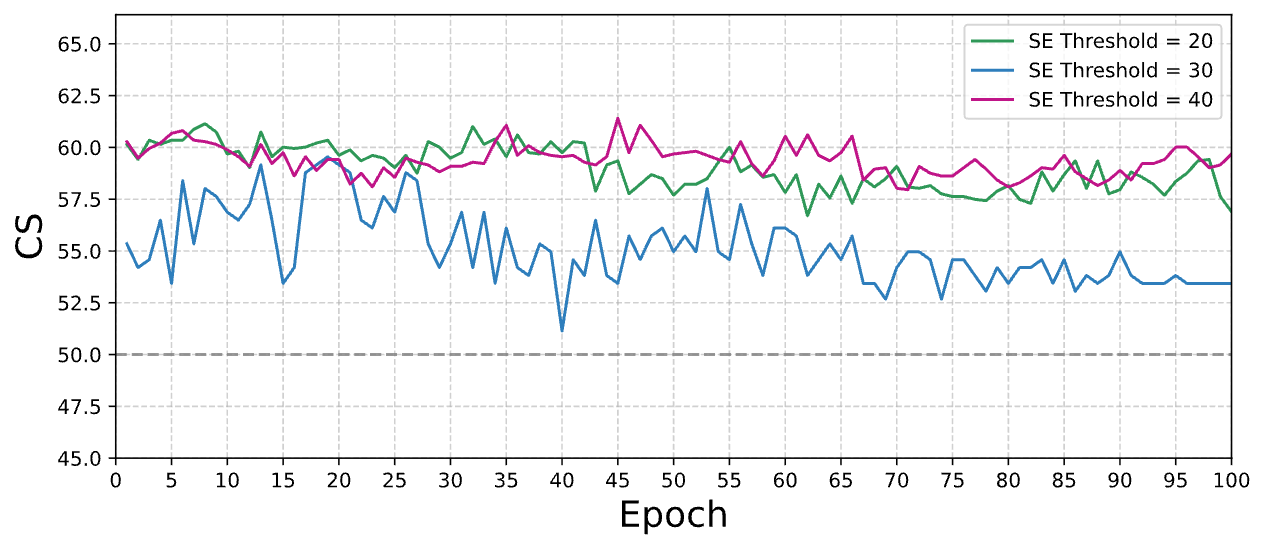}
    \caption{BERT CS Score for SE thresholds 20, 30 and 40.}
    \label{bert-203040-cs}
\end{minipage}

% \end{center}

% \begin{center}

\begin{minipage}{0.48\linewidth}
    \centering
    \includegraphics[width=1\linewidth, height=3cm,
  keepaspectratio=false]{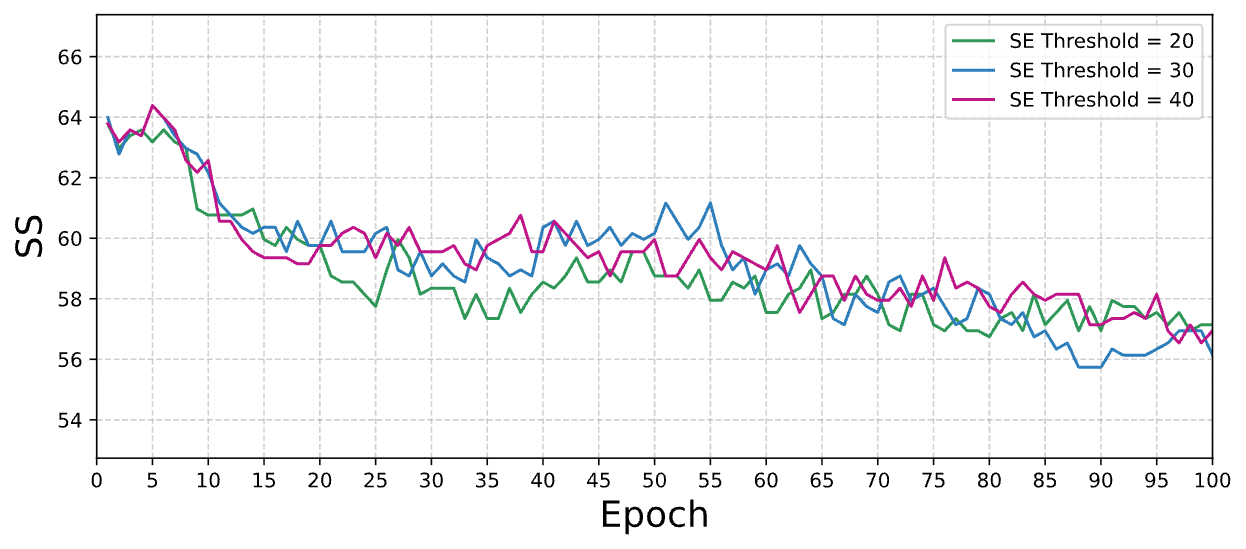}
    \caption{GPT-2 SS Score SE thresholds 20, 30 and 40.}
    \label{gpt2-203040-ss}
\end{minipage}
\hfill
\begin{minipage}{0.48\linewidth} % Keep the same width as before
    \centering
    \includegraphics[width=1\linewidth, height=3cm,
  keepaspectratio=false]{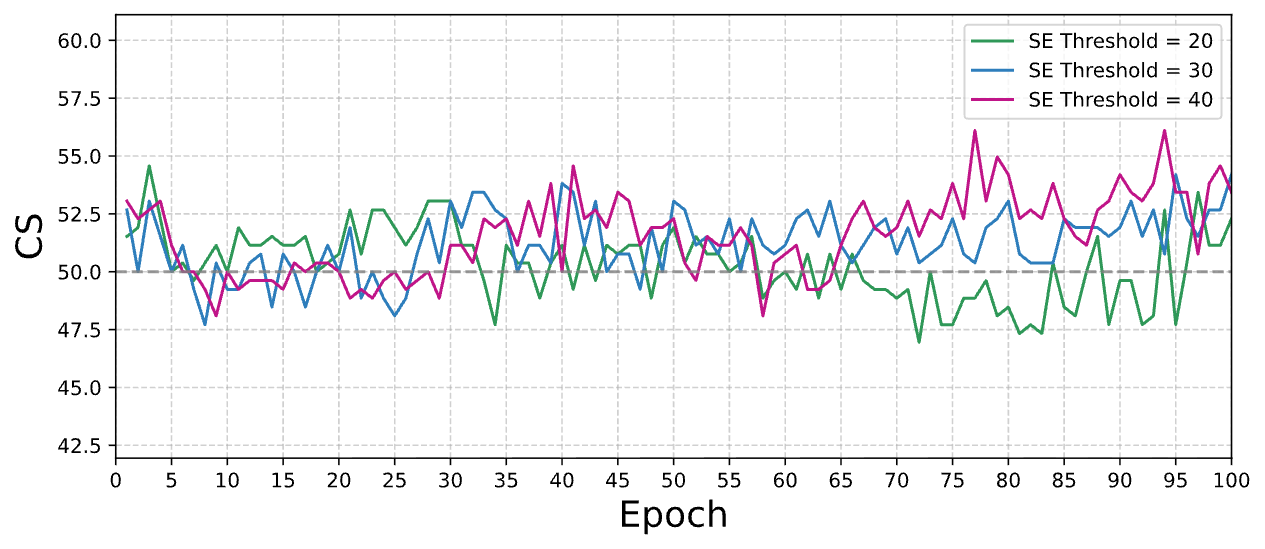}
    \caption{GPT-2 CS Score for SE thresholds 20, 30 and 40.}
    \label{gpt2-203040-cs}
\end{minipage}

% \end{figure*}

% \begin{figure*}[t]

\centering
% \begin{center}
\begin{minipage}{0.48\linewidth}
    \centering
    \includegraphics[width=1\linewidth, height=3cm,
  keepaspectratio=false]{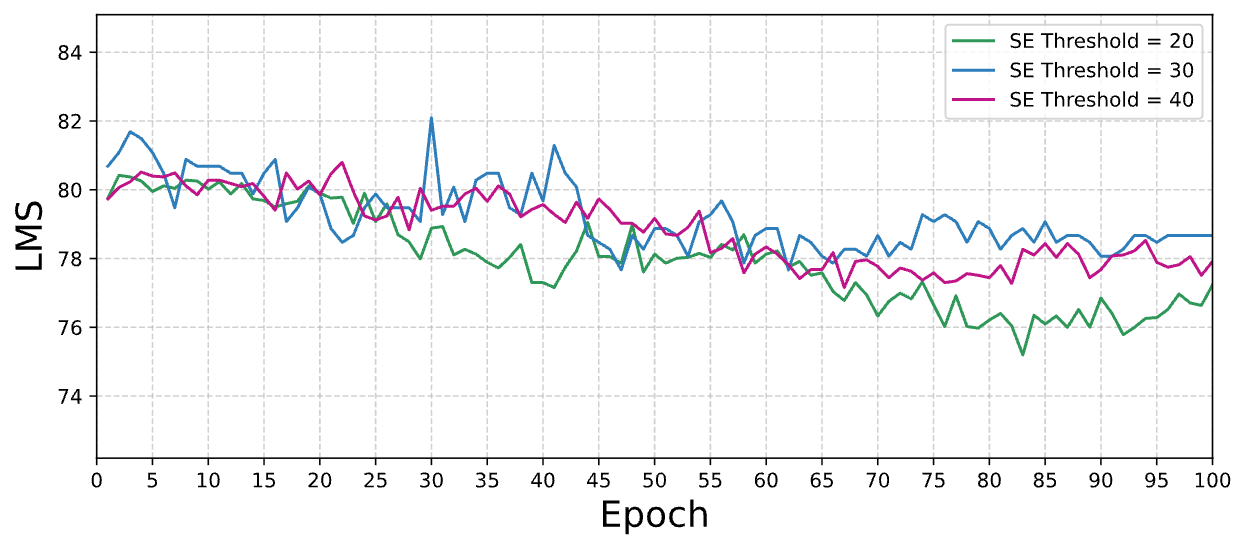}
    \caption{BERT LMS Score for SE thresholds 20, 30 and 40.}
    \label{bert-203040-lms}
\end{minipage}
\hfill
\begin{minipage}{0.48\linewidth} % Keep the same width as before
    \centering
    \includegraphics[width=1\linewidth, height=3cm,
  keepaspectratio=false]{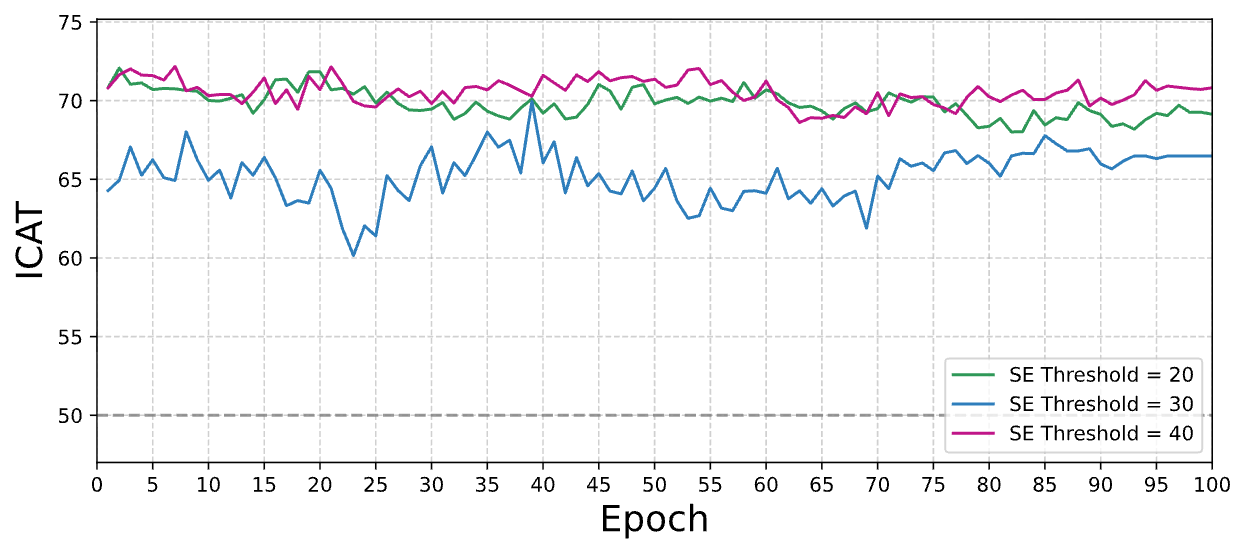}
    \caption{BERT ICAT Score for SE thresholds 20, 30 and 40.}
    \label{bert-203040-icat}
\end{minipage}
% \end{center}

% \begin{center}
\begin{minipage}{0.48\linewidth}
    \centering
    \includegraphics[width=1\linewidth, height=3cm,
  keepaspectratio=false]{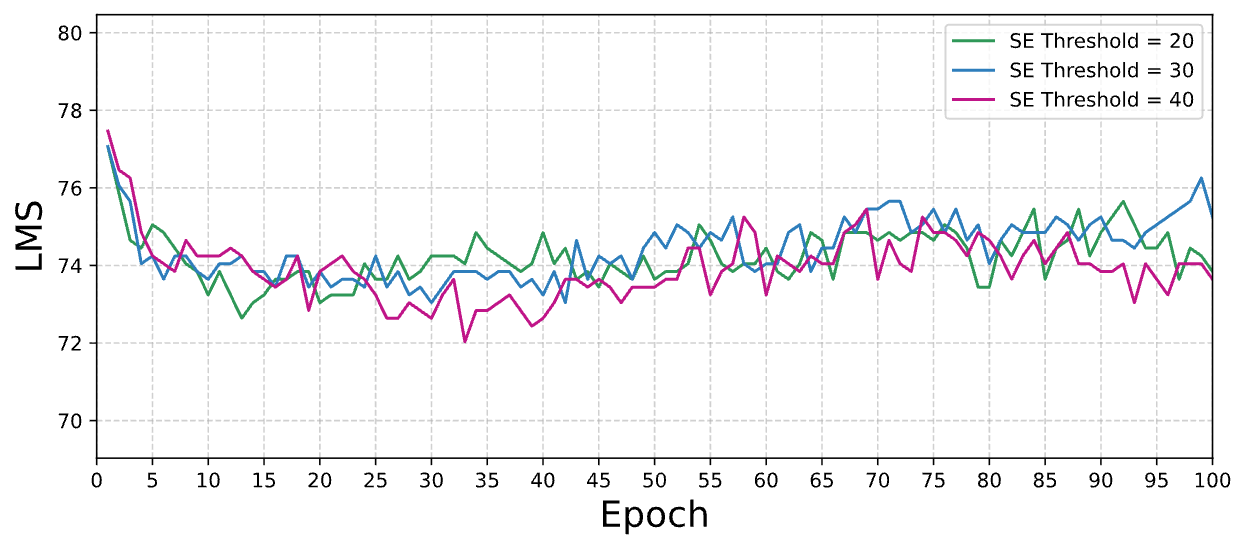}
    \caption{GPT-2 LMS Score for SE thresholds 20, 30 and 40.}
    \label{gpt2-203040-lms}
\end{minipage}
\hfill
\begin{minipage}{0.48\linewidth} % Keep the same width as before
    \centering
    \includegraphics[width=1\linewidth, height=3cm,
  keepaspectratio=false]{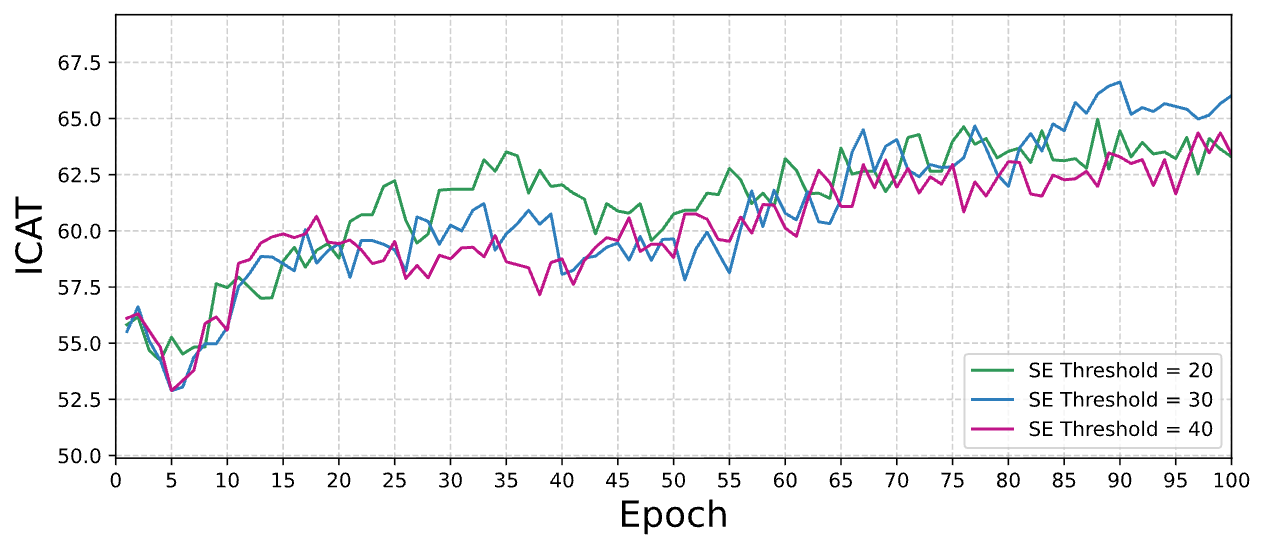}
    \caption{GPT-2 ICAT Score for SE thresholds 20, 30 and 40.}
    \label{gpt2-203040-icat}
\end{minipage}
% \end{center}

\end{figure*}

\vspace{0.3em}
\textbf{Impact on debiasing performance: }Figures \ref{bert-203040-ss}–\ref{gpt2-203040-cs} illustrate the effect of semantic entropy thresholds on intrinsic bias metrics such as Stereotype Score (SS) and CrowS-Pairs Score (CS) for both BERT and GPT-2. At a 20\% threshold (i.e., more lenient filtering), the model retains more counterfactuals, including those with moderate uncertainty. This sometimes leads to insufficient bias removal, particularly visible in the slightly higher CS values in Fig. \ref{bert-203040-cs}. Conversely, at a 40\% threshold (i.e., more aggressive filtering), while bias mitigation improves initially due to the exclusion of noisier samples, over-filtering may reduce the diversity of the counterfactual corpus, potentially limiting the coverage of gender-related variations and reducing generalization. Overall, the 30\% threshold achieves the best balance: it significantly reduces gender bias and outperforms CDA while avoiding the potential drawbacks of both under- and over-filtering as observed in Fig. \ref{bert-203040-cs}, \ref{gpt2-203040-ss} and \ref{gpt2-203040-cs}. Hence, we choose 30\% as the optimal setting for entropy-based filtering in the main experiments. However, the higher SS scores in Fig. \ref{bert-203040-ss} indicate the need for a more fine-tuned evaluation of this threshold to arrive at a balanced and optimal value.

% \vspace{0.3\baselineskip}
\par\vspace{0.3em}
\textbf{Impact on Language Modeling Performance: }Figures \ref{bert-203040-lms}–\ref{gpt2-203040-icat} report the Language Modeling Score (LMS) and ICAT score under different entropy thresholds-20, 30 and 40. For both BERT and GPT-2, a 20\% threshold tends to retain too many noisy samples, leading to slightly reduced fluency and coherence, as seen in lower LMS scores in Fig. \ref{bert-203040-lms}. On the other hand, a 40\% threshold, while helping eliminate uncertain samples, may discard too many useful and contextually rich augmentations. In contrast, the 30\% threshold yields higher LMS and ICAT scores, indicating better preservation of language fluency and semantic understanding. These results reinforce that filtering at this level helps strike an optimal trade-off between reducing uncertainty and retaining the linguistic richness of the training corpus. However, these thresholds can be further refined to reach a more balanced threshold.

\end{document}